\documentclass{article} % For LaTeX2e
\usepackage{colm2024_conference}

\usepackage{microtype}
\usepackage{hyperref}
\usepackage{url}
\usepackage{booktabs}

\usepackage{inconsolata}
\usepackage{graphicx}
\usepackage{algorithm}
\usepackage{algorithmic}
\usepackage{amsmath}
\usepackage{enumitem}
\usepackage{cuted}
\usepackage{xcolor}
\usepackage{subcaption}
\usepackage{booktabs}
\usepackage{afterpage}
\usepackage{multirow}
\usepackage{pifont}
\usepackage{amsfonts}
\usepackage{hyperref}
\usepackage{wrapfig}
\usepackage{caption}
\usepackage{amsmath}
\usepackage{array}

\newcommand{\xmark}{\ding{55}}
\newcommand{\graycomment}[1]{\textcolor{gray}{/* #1 */}}

\usepackage{listings}
\title{\centerline{Interpretable-by-Design Text Understanding}
\centerline{with Iteratively Generated Concept Bottleneck}}

\author{\centerline{Josh Magnus Ludan, Qing Lyu, Yue Yang, Liam Dugan,} \vspace{0.1cm} \\ \centerline{\textbf{Mark Yatskar}, \textbf{Chris Callison-Burch}} \vspace{0.15cm} \\ \centerline{University of Pennsylvania} \vspace{0.05cm} \\
\centerline{\small{\{jludan, lyuqing, yueyang1, ldugan, myatskar, ccb\}@seas.upenn.edu}}}

\colmfinalcopy % Uncomment for camera-ready version, but NOT for submission.
\begin{document}

\maketitle

\begin{abstract}
Black-box deep neural networks excel in text classification, yet their application in high-stakes domains is hindered by their lack of interpretability. To address this, we propose \textbf{T}ext \textbf{B}ottleneck \textbf{M}odels (\textbf{TBM}), an intrinsically interpretable text classification framework that offers both global and local explanations. Rather than directly predicting the output label, TBM predicts categorical values for a sparse set of salient concepts and uses a linear layer over those concept values to produce the final prediction. These concepts can be automatically discovered and measured by a Large Language Model (LLM) without the need for human curation. Experiments on 12 diverse text understanding datasets demonstrate that TBM can rival the performance of black-box baselines such as few-shot GPT-4 and finetuned DeBERTa while falling short against finetuned GPT-3.5. Comprehensive human evaluation validates that TBM can generate high-quality concepts relevant to the task, and the concept measurement aligns well with human judgments, suggesting that the predictions made by TBMs are interpretable.
Overall, our findings suggest that TBM is a promising new framework that enhances interpretability with minimal performance tradeoffs.
% particularly for sentiment classification. 
\footnote{Code is available at \href{https://github.com/JMRLudan/TBM}{github.com/JMRLudan/TBM}.}
\end{abstract}

\section{Introduction}
Interpretability has become a critical aspect of deep learning systems, especially in high-stakes domains such as law, finance, and medicine, where understanding and analyzing model behavior is crucial \citep{bhatt2020explainable, dwivedi2023explainable}.
A promising line of work focuses on ``self-interpretable'' models, which provide built-in explanations along with their predictions \citep{du2019techniques, linardatos2020explainable}. These model-provided explanations can come in various forms: token-level importance scores, influential training examples, or even free text. However, these types of explanations oftentimes provide only \textit{local} justification for individual predictions and fail to offer \textit{global} insights into the overarching principles that guide model behavior \citep{bhatt2020explainable}.

An alternative form of explanation that addresses this issue is \textit{concept}-based explanations \citep{madsen2022post}. A concept is a high-level feature representing some aspect of the input text, such as ``food quality'' for a restaurant review. Concept-based explanations can provide both global and local insights by identifying important concepts across the dataset and localizing how these concepts relate to each individual prediction. However, concept-based approaches typically involve extensive human labor since they require experts to curate a set of concepts for each new task, and the concept values need to be further annotated on each training example \citep{abraham2022cebab}. Additionally, current approaches include hundreds or even thousands of concepts in their explanations \citep{rajagopal-etal-2021-selfexplain}. With such large concept spaces, it remains difficult to draw useful takeaways on the global behavior of the model \citep{ramaswamy2022overlooked}.

In this work, we propose \textbf{T}extual \textbf{B}ottleneck \textbf{M}odel (\textbf{TBM}), an extension of Concept Bottleneck Models (CBM) \citep{kohbottlenecks} to text classification and regression tasks. Our system has three fully automated modules: Concept Generation, Concept Measurement, and a Prediction Layer, as shown in Figure~\ref{fig:interpretability_demo}. Given a dataset of input texts (e.g., restaurant reviews), the Concept Generation module iteratively discovers a sparse set of concepts (e.g., ``Menu Variety'') that help discriminate between texts with different output labels. The Concept Measurement module then determines the value of each concept (e.g., ``wide variety'') for a text as a numerical score (e.g., +1). Finally, these concept scores are aggregated into the final prediction by a white-box Prediction Layer (e.g., a linear layer).

\begin{wrapfigure}{r}{0.5\textwidth}
\vspace{-.5cm}
  \begin{center}
    \includegraphics[width=0.48\textwidth]{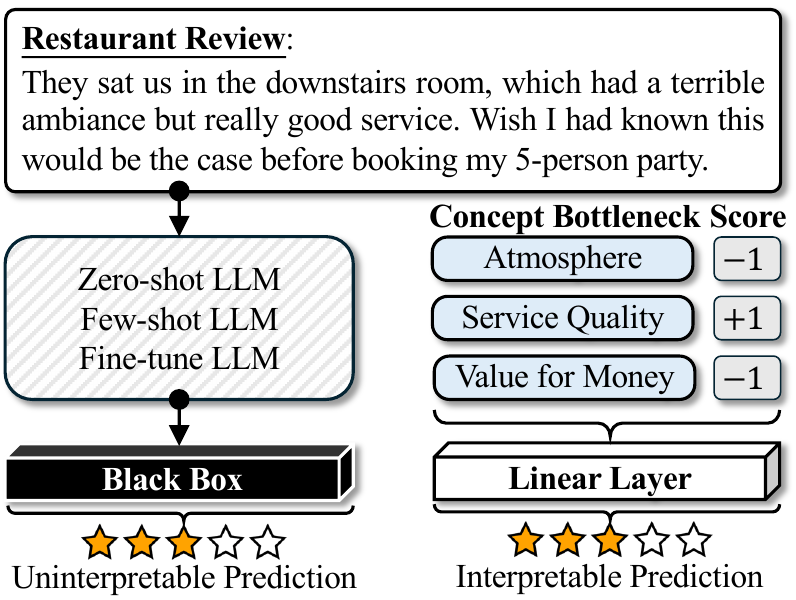}
  \end{center}
  \vspace{-.48cm}
  \caption{Unlike end-to-end black-box language models (left), Text Bottleneck Models (right) first discover and measure a set of human-readable concepts and then predict the final label with an interpretable linear layer.}
\label{fig: intro example}
\vspace{-.5cm}
\end{wrapfigure}

We evaluate our system on 12 diverse datasets, including sentiment classification, natural language inference, intent detection, and topic classification. TBM performs competitively with strong black-box baselines, including few-shot GPT-3.5 and finetuned BERT, but lags behind state-of-the-art models like finetuned GPT-3.5. In particular, TBM is competitive for sentiment analysis, though there is room for improvement in specialized domains like news and science. 

To understand where the error comes from, we manually evaluate each module and find that the Concept Generation module can consistently generate high-quality concepts but can occasionally struggle with redundancy and leakage. The Concept Measurement module is found to score the majority of concepts in sentiment analysis with high accuracy, whereas those in fake news detection are harder to measure, which might be a reason behind the performance difference in these domains. Finally, the concept learning curves make it transparent what concepts are learned over time and their relative impact, which can offer valuable insights for model understanding and debugging.

In summary, our contributions in this paper are threefold: First, we introduce TBM, a text classification framework that provides both global and local interpretability, by automatically constructing sparse concept bottlenecks using LLMs without any human effort. Second, we demonstrate that, on average, TBM performs competitively with 
strong, but not state-of-the-art, black-box baselines across 12 diverse datasets. Third, we provide an in-depth human evaluation and analysis of each module in the TBM.
\section{Related Work}

\textbf{Self-interpretable NLP models} aim to provide a built-in explanation along with the prediction without relying on post-hoc explanation methods. They offer diverse forms of explanation. Token-based explanations, such as rationales \citep{zaidan-etal-2007-using,lei2016rationalizing, bastings2019interpretable}, provide a span of important tokens that are minimally sufficient for the prediction. Example-based explanations \citep{han-etal-2020-explaining, das2022prototex} identify the most similar examples within the training set relative to the examples for inference. Free-text explanations, such as those in \citep{camburu2018snli,nye2021show, wei2022chain}, generate a free-form justification in natural language for the prediction. We note that these only provide local interpretability, and our approach differs in that it provides both local and global insights into model behavior owing to concept-based explanations.

\textbf{Concept Bottleneck Models} were first introduced by \citep{kohbottlenecks} for vision tasks. Their work tasked experts with manually crafting a set of human-interpretable concepts that became the only input for a classifier model. Stakeholders could then intervene and correct these concepts, allowing easier model behavior analysis. \citep{collins2023human} describe several problems with CBMs, such as information leakage \citep{mahinpei2021promises} and having too many concepts \citep{ramaswamy2022overlooked}. Information leakage causes the concept bottleneck to be \textit{unfaithful} \citep{jacovi-goldberg-2020-towards, lyu2022towards} by having the labeling task as a concept. Having too many concepts causes information overload for the user, preventing them from developing a general understanding of model behavior. We note that these problems can also exist in the text domain, so we carefully evaluate them in our manual analysis.\footnote{See Sec~\ref{sec: Concept Generation Module Evaluation} for ``Redundancy'' and ``Leakage'' evaluation.}
Previous work in computer vision has also used LLMs to automate this process for image classification \citep{yang2023language, pratt2023does} to reduce the cost of concept generation. Our work extends this method to the text domain, with additionally introduced benefits such as sparsity.

\textbf{Concept-based explanations in NLP} can be broadly categorized into two lines of work. The first focuses on mechanistic interpretability, analyzing \textit{what} latent concepts are represented by different neurons in pre-trained LMs \citep{sheng2020investigating, bills2023language, vig2020investigating}. The second focuses on explaining \textit{why} models make certain decisions, providing explicit concepts as supporting evidence for predictions. Our work belongs to the second category. 
Within this category, SELF-EXPLAIN \citep{rajagopal-etal-2021-selfexplain} is an explainable framework that jointly predicts the final label and identifies globally similar concepts from the training set and locally relevant concepts from the current example. Notably, their approach has no bottleneck structure, which makes information leakage easier. Also, they define each \textit{phrase} (e.g., ``for days'', ``the lack of'', etc.) in each example as a concept, resulting in an enormous concept space of hundreds of thousands of phrases. By contrast, our concepts are high-level, categorical \textit{features}, resulting in a sparse space of $\leq 30$ concepts for each dataset, making it easier to draw useful takeaways. Another representative work \citep{wu2023causal} trains a Causal Proxy Model that mimics the behavior of a black-box model using human-annotated counterfactual data. Our definition of concepts is consistent with theirs but does not require expert data curation.
\section{Method}
\begin{figure*}
\begin{center}
\includegraphics[width=\textwidth]{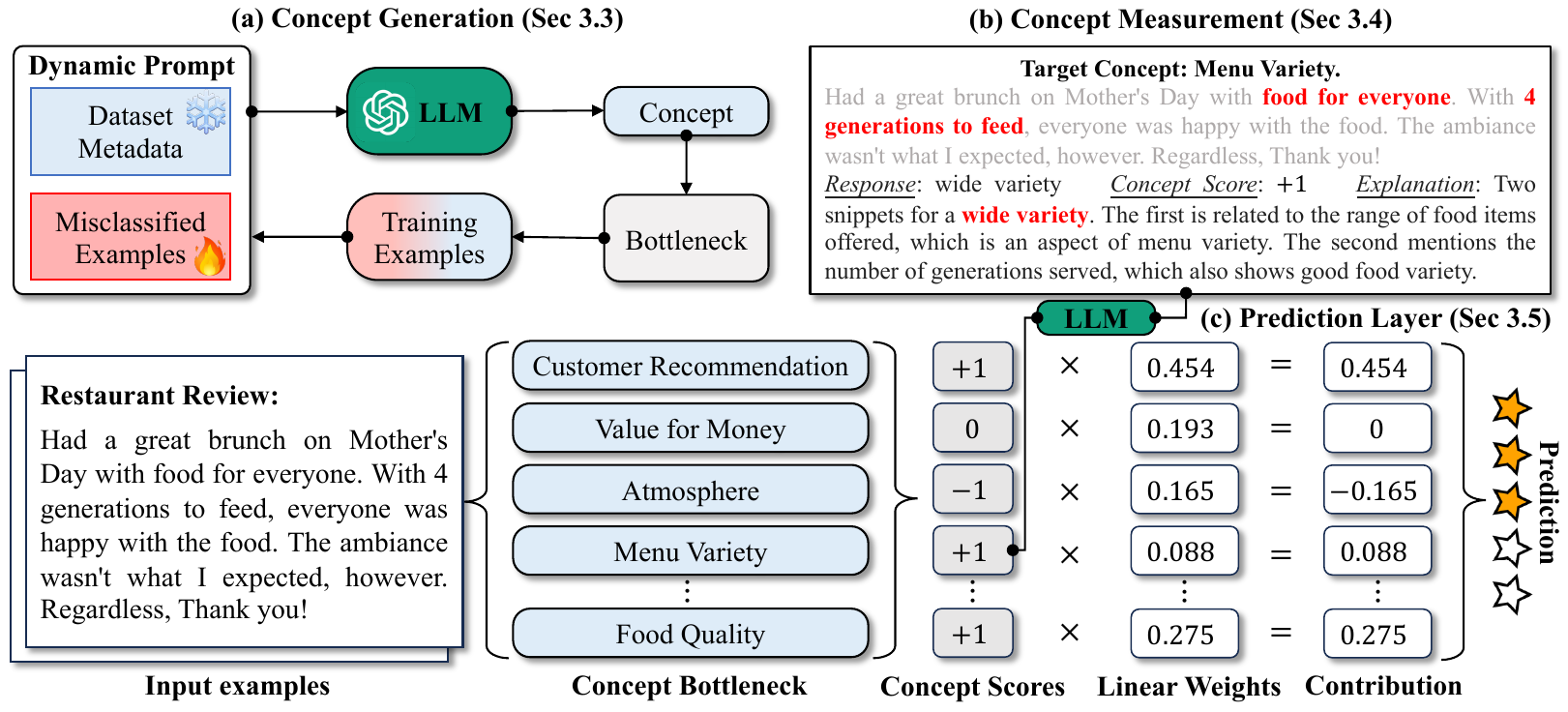}
\end{center}
\vspace{-.4cm}
\caption{Demonstration of the \textbf{T}extual \textbf{B}ottleneck \textbf{M}odel (\textbf{TBM}) with an example from the CEBaB \citep{abraham2022cebab} dataset. Given an input example (restaurant review), during (a) \textbf{Concept Generation} (Sec \ref{sec: concept generation}), it iteratively discovers new concepts (e.g., ``Menu Variety''). (b) \textbf{Concept Measurement} (Sec \ref{sec: concept measurement}) measures the value of concepts by identifying relevant snippets (e.g., ``food for everyone'') and providing a numerical concept score (e.g., $+1$). Finally, the (c) \textbf{Prediction Layer} (\ref{sec: linear head}) aggregates all concept scores for the input and learns their relative weights to make the final prediction of the task label.}
\label{fig:interpretability_demo}
\vspace{-.4cm}
\end{figure*}

Figure~\ref{fig:interpretability_demo} provides an overview of our system. It consists of three components: (1) \textbf{Concept Generation} (Sec \ref{sec: concept generation}), which iteratively discovers new concepts using misclassified examples; (2) \textbf{Concept Measurement} (Sec \ref{sec: concept measurement}), which measures the concept scores for each example; and the (3) \textbf{Prediction Layer} (Sec \ref{sec: linear head}), which predicts the output label with only the concept scores as input. The first two modules are implemented by prompting an LLM,\footnote{See Appendix~\ref{sec: appendix prompt} for all relevant prompts.} and the last module is implemented as training a linear layer.

\subsection{Method Formulation}
We describe the structure of TBM as follows:\footnote{Note that this is a generic structure of TBM independent of implementation.} Given a text classification or regression dataset with a training set $\mathcal{D}_\text{train}$ and a test set $\mathcal{D}_\text{test}$, each instance can be denoted as a text-label pair $(t, y)$. During training, we generate a set of $N$ concepts $C = \left\{c_1, c_2, \ldots c_N\right\}$ using $\mathcal{D}_\text{train}$, where each concept $c_i$ is a categorical feature (e.g., ``menu variety'') with multiple possible values (e.g., \texttt{high}, \texttt{low}, \texttt{mixed} or \texttt{unmentioned}). For each text $t_\text{train}$, we measure the values of all concepts as a list of numerical scores, $\left[s\left(t_\text{train}, c_i\right) | c_i \in C\right]$ (e.g., $+1$, $-1$, $0$). The sign of the score represents the polarity of a concept in the text, i.e., a positive/negative score indicates that the concept is positively/negatively reflected, and a zero score represents the uncertainty or absence of the concept. The importance of a concept is reflected by the magnitude of the score, with a larger magnitude indicating higher intensity. These concept scores are then used as the only input to train a white-box prediction layer to predict the label $y_\text{train}$.
%in a separate test set $\mathcal{D}_\text{test}$. 
During inference, given a new input text $t_\text{test} \in \mathcal{D}_\text{test}$, we measure the score of each concept in the generated concept set $\left[s\left(t_\text{test}, c_i\right) | c_i \in C\right]$, and use the trained prediction layer to predict the final label $y_\text{test}$.

In the following sections, using Figure~\ref{fig:interpretability_demo} as a running example, we describe our specific implementation of each TBM module in terms of how concepts are represented, generated, and measured and how these concept measurements are turned into predictions.

\begin{wraptable}{r}{7.5cm}
\tiny
\centering
\vspace{-.4cm}
\begin{tabular}{p{0.25\linewidth} p{0.63\linewidth}}
\toprule
\textbf{Key} & \textbf{Value} \\
\midrule
Concept Name & Build Quality \\
\midrule
Concept Description & Build quality refers to the craftsmanship, durability, and overall construction of a product. This concept encompasses various aspects such as the materials used, design, manufacturing techniques, and attention to detail. \\
\midrule
Concept Question & What does the review say about the build quality of the product? \\
\midrule
Possible Responses & Positive, Negative, Uncertain, Not applicable \\
\midrule
Response Guide & \textbf{Positive:} The review mentions positive aspects such as being well-made, sturdy, durable, use of high-quality materials, excellent craftsmanship, etc.\\
& \textbf{Negative:} The review mentions negative aspects such as poor construction, flimsiness, use of cheap materials, bad design, being easily breakable, etc.\\
& \textbf{Uncertain:} The review does not clearly mention the build quality, provides ambiguous or vague information, or mentions both positive and negative aspects.\\
& \textbf{Not applicable:} The review does not mention the build quality of the product at all. \\
\midrule
Response Mapping & Positive: $+1$, Negative: $-1$, \\ & Uncertain: $0$, Not applicable: $0$ \\
\bottomrule
\end{tabular}
\caption{JSON Representation for the concept ``Build Quality'' for a hypothetical product review dataset included in the Concept Generation prompt as an in-context example.}
\label{tab:json_representation}
\vspace{-.8cm}
\end{wraptable}
\subsection{Concept Representations} 
\label{sec:concept representation}

Each concept consists of the following components, represented as a JSON object in our prompts: (1) \textbf{Concept Name}: The name of the concept; (2) \textbf{Concept Description}: A description of the concept and the factors relevant to measuring it; (3) \textbf{Concept Question}: The question we use to measure the concept value; (4) \textbf{Possible Responses}: The set of possible responses to the concept question; (5) \textbf{Response Guide}: A list of criteria for possible responses, to guide the process of answering the concept question; (6) \textbf{Response Mapping}: A dictionary mapping each possible response to a numerical score.

Table \ref{tab:json_representation} shows an example concept ``Build Quality'' for a product review dataset. The concept question and response guide are important during the Concept Measurement stage.

\subsection{Concept Generation} \label{sec: concept generation}
  \begin{wrapfigure}{R}{0.5\textwidth}
    \vspace{-.85cm}
    
    \begin{minipage}{0.5\textwidth}
      \begin{algorithm}[H]
      \captionsetup{font=small}
      \small
        \caption{Iterative Concept Generation}
        \begin{algorithmic}
            \STATE $ \mathcal{D}_\text{train} $ $ \gets $ training samples
            \STATE $ C $ $ \gets $ $\left[\right]$, list of concepts
            \STATE \textbf{initialize} TBM \textbf{with} $ \mathcal{D}_\text{train} $, $ C = []$.
            \smallskip
            \FOR{$i = 1$ to $N$}
            \STATE \graycomment{\textit{Identify misclassified examples.}}
            \STATE $ \mathcal{D}_\text{mis} $ $ \gets $ $ \left\{\left(t, y\right) \in \mathcal{D}_\text{train} \,|\, \text{TBM}(t) \neq y\right\}$
            \smallskip
            \STATE \graycomment{\textit{Prompt with misclassified examples to generate a new concept}}
            \STATE $ c $ $ \gets $ \textbf{GeneratePrompt}$\left( \mathcal{D}_\text{mis}, C\right)$
            \smallskip
            \STATE \graycomment{\textit{Refine the generated concept.}}
            \STATE $ c' $ $ \gets $ \textbf{RefinePrompt}$\left(c\right)$
            \smallskip
            \STATE \graycomment{\textit{Train TBM with updated concepts.}}
            \STATE TBM$'$ $ \gets $ \textbf{train} TBM \textbf{with} $ \mathcal{D}_\text{train} $, $C$.\text{append}($c'$)
            \smallskip
            \STATE \graycomment{\textit{Admit concept if model is improved.}}
            \IF{TBM$'$.score $-$ TBM.score $> \gamma$}
            \STATE $ C $ $ \gets $ $ C.\text{append}(c') $
            \STATE TBM $ \gets $ TBM$'$
            \ENDIF
            \ENDFOR
        \end{algorithmic}
        \label{fig:algorithm}
      \end{algorithm}
    \end{minipage}
    \vspace{-1.4cm}
  \end{wrapfigure}
At a high level, we generate concepts by prompting an LLM to discover new concepts that help discriminate between misclassified examples iteratively. As outlined by Algorithm~\ref{fig:algorithm}, given the training set (e.g., restaurant reviews), we initialize the TBM with an empty concept set $C$. In each iteration, to generate a new concept $c$, we first identify training examples with similar representations in the existing concept space but with a high prediction error under the current Prediction Layer. For example, if the current concept space $C$ contains only ``Atmosphere'' ($c_1$) and ``Food Quality'' ($c_2$), then the two reviews \textit{Great food and ambiance, but quite limited choices on the menu} (3-star) and \textit{Food, atmosphere, variety of choices... everything was excellent!} (5-star) will both be represented as $[c_1=+1, c_2=+1]$. However, a new concept, ``Menu Variety'', can help differentiate between them.
Therefore, we construct the concept generation prompt (GeneratePrompt) using the dataset metadata (description and labeling scheme) and these hard examples as in-context exemplars to encourage the generation of a new discriminative concept. We also include the list of previously generated concepts in this prompt to reduce concept duplication.

Taking GeneratePrompt as input, the LM generates a new candidate concept $c$, which will then be refined through RefinePrompt. RefinePrompt contains a few examples of problematic concepts,  such as those with ambiguous questions and invalid JSON formatting, and how they are fixed. The resulting refined concept, $c'$, and the existing $C$ are used to train a new Prediction Layer to create a new candidate TBM$'$. If TBM$'$ outperforms existing TBM on a random subset of $ \mathcal{D}_\text{train} $ by some threshold $\gamma$, it is retained, otherwise omitted. The above procedure is iteratively executed for $ N $ cycles, resulting in a final concept set $C$.\footnote{We set $\gamma= $ and $N=30$ for all experiments in this paper and the size of the $ \mathcal{D}_\text{train} $ subset to 100.}

\subsection{Concept Measurement} \label{sec: concept measurement}

With the generated concept set $C$, the Concept Measurement module determines the scores $\left[s\left(t, c_i\right) | c_i \in C\right]$ for any given text $ t $. To measure a concept, we prompt an LLM in a zero-shot fashion to answer the concept question associated with that concept, using the concept description and response guide as context (see Sec~\ref{sec:concept representation}). For instance, the concept ``Menu Variety'' in Figure~\ref{fig: intro example}. Given a restaurant review, the concept question prompts, ``How does the review describe the variety and originality of the menu?'' The possible answers could be ``wide variety'', ``low variety'', ``uncertain'', or ``not applicable''. The response given by the LLM is then converted into a numerical concept score using the concept's response mapping (+1 for Positive, -1 for Negative, 0 otherwise). In addition to the categorical answer, the prompt instructs the LLM to provide relevant snippets in the input text as supporting evidence. For example, ``food for everyone, with 4 generations to feed'' is the supporting snippet for ``wide variety''.

\subsection{Prediction Layer} \label{sec: linear head}

To combine the concept scores $\left[s\left(t, c_i\right) | c_i \in C\right]$ into a final prediction $y$, we train a Prediction Layer on $\mathcal{D}_\text{train}$, using linear regression for regression tasks and logistic regression for classification tasks.\footnote{We note that any interpretable classifier that operates on numerical concept scores, such as decision trees, can be used for the final classification.} It learns a weight associated with each concept using $y$ as the supervision signal. For a new input example at inference time, its measured concept scores are multiplied by their weights and summed into the final prediction logit. For example, in Figure~\ref{fig:interpretability_demo}, across the dataset, ``Customer Recommendation'' and ``Food Quality'' are the most important concepts, while ``Menu Variety'' is less crucial. On the given review, ``Customer Recommendation'' and ``Menu Variety'' are positively scored, but ``Atmosphere'' and ``Value for Money'' are negatively scored. Their weighted sum results in a final prediction of 3 stars. 

Finally, the concept weights provide a \textit{global} explanation for their relative importance across the dataset, and the concept scores and supporting snippets provide a \textit{local} explanation for the decision on each example.

% The example in Figure~\ref{fig: intro example} shows how the concept scores multiply with their learned linear weights to calculate the contributions in the final prediction.

% \input{tables/main_results_figure}
% \input{tables/analysis_results}
\section{Experimental Setup}
\textbf{Datasets.} We evaluate a total of 12 datasets. Six of them involve sentiment understanding, including \textbf{Rotten Tomatoes} \citep{Pang+Lee:05a}, \textbf{Amazon reviews} \citep{mcauley2013hidden}, \textbf{Poem Sentiment} \citep{sheng2020investigating}, \textbf{CEBaB} restaurant reviews \citep{abraham2022cebab}, \textbf{Yelp reviews} \citep{zhang2015character}, and \textbf{Hate Speech Detection} \citep{kennedy2020constructing}. The rest involve multiple diverse tasks, including Natural Language Inference (\textbf{SNLI} \citep{snli:emnlp2015}), Topic Classification (\textbf{AG News} \citep{Gulli}, \textbf{Patent Classification} \citep{sharma2019bigpatent}), Intent Detection (\textbf{SciCite} \citep{cohan-etal-2019-structural} ), and news analysis (\textbf{Fake News Detection} \citep{perez-rosas-etal-2018-automatic}, \textbf{News Partisanship Classification} \citep{kiesel-etal-2019-semeval}).
% Stanford Natural Language Inference (\textbf{SNLI}) \citep{snli:emnlp2015}, \textbf{Fake News Detection} \citep{perez-rosas-etal-2018-automatic}, \textbf{News Partisanship Classification} \citep{kiesel-etal-2019-semeval}, \textbf{Citation Intent Detection} \citep{cohan-etal-2019-structural}, \textbf{AG News} \citep{Gulli}, \textbf{Patent Classification} \citep{sharma2019bigpatent}.
More details of these datasets can be found in Appendix~\ref{sec: Appendix Datasets}. Three of these datasets involve a regression task (CEBaB, Yelp, and Hate Speech), while the rest involve classification. With few noted exceptions, we train TBM using 250 examples and test on 250 examples for each dataset considering the expense from API queries.

\textbf{Baselines.} For comparison, we choose the following 5 baselines: (1)\textbf{DeBERTa} \citep{he2021deberta}: We finetune a \href{https://huggingface.co/microsoft/deberta-base}{DeBERTa-base} classifier for three epochs. (2) \textbf{BERT} \citep{DBLP:journals/corr/abs-1810-04805}: We finetune a \href{https://huggingface.co/bert-base-uncased}{bert-base-uncased} classifier for three epochs. (3) \textbf{Naive Bayes} \citep{mccallum1998comparison}: We fit a Naive Bayes classifier on top of the TF-IDF matrix of the texts as an interpretable baseline. (4) \textbf{GPT-4 (10-shot)}: We use up to 10 examples\footnote{We reduce the number of examples if the maximum context length is reached.} with labels to prompt GPT-4 (\texttt{gpt-4-0613}).
(5) \textbf{GPT-3.5-turbo (finetune)} \citep{Peng_Wu_Allard_Kilpatrick_Heidel_2023}: We finetune a GPT-3.5-turbo model (\texttt{gpt-3.5-turbo-1106}) for three epochs.

\textbf{Evaluation Metrics.} We evaluate TBMs in three ways. First, we compute the end-to-end performance (Mean Squared Error (MSE) for regression and accuracy for classification) compared to the above baselines. Next, we evaluate the Concept Generation and Concept Measurement modules using human annotation (see metrics in Sec~\ref{sec: results}).

\textbf{Implementation Details.} We use GPT-4 (\texttt{GPT-4-0613}) as the underlying LLM for Concept Generation and Concept Measurement. We use Scikit-learn \citep{scikit-learn} for training linear regression (regression tasks) and logistic regression (classification tasks).

\begin{table*}[!t]
\centering
\small
% \resizebox{\textwidth}{!}{%
\begin{tabular}{l|cccc|cccc}
\toprule
\multirow{2}{*}{\begin{tabular}[c]{@{}c@{}}\textbf{Model}  (Interpretable)\end{tabular}} & \textbf{Rotten} & \textbf{Amaz} & \textbf{Poem} & \textbf{Avg} & \textbf{CEBaB} & \textbf{Yelp} & \textbf{Hate} & \textbf{Avg}   \\ %\cline{2-10}
& Acc$\uparrow$ & Acc$\uparrow$ & Acc$\uparrow$ & Acc$\uparrow$ & MSE$\downarrow$ & MSE$\downarrow$ & MSE$\downarrow$ & MSE$\downarrow$  \\\midrule
BERT-base (\xmark) &  0.788 & 0.872 & 0.712 & 0.791 & 0.567 & 0.935 & 1.868 & 1.123 \\
DeBERTa-base (\xmark) &  0.824 & 0.924 & 0.728 & 0.825 & 0.346 & 0.539 & 1.712 & 0.866\\
% GPT-3.5 (5-shot) (\xmark) &  \\
GPT-3.5 (fine-tuned) (\xmark) & \textbf{0.916} & 0.964 & \textbf{0.820} & \textbf{0.900} & \underline{0.300} & \textbf{0.400} & \textbf{1.079} & \textbf{0.593}\\  
GPT-4 (10-shot) (\xmark) & \underline{0.912} & \textbf{0.980} & 0.628 & 0.840 & \textbf{0.272} & \underline{0.412} & 1.666 & 0.783\\  
\midrule
Naive Bayes (\checkmark) &  0.640 & 0.684 & 0.604 & 0.643 & 1.173 & 1.380 & 2.789 & 1.781 \\
TBM (Ours) (\checkmark)  & \textbf{0.916} & \underline{0.972} & \underline{0.800} & \underline{0.896} & 0.367 & 0.467 & \underline{1.179} & \underline{0.671}\\
% $\Delta$ (\%) vs. few-shot \\
% $\Delta$ (\%) vs. fine-tune \\
\bottomrule
\end{tabular}
% }
\caption{Model performance on 6 \textbf{sentiment} datasets with 3 for regression and 3 for classification. \xmark~and \checkmark~denote whether the model is interpretable or not. For each dataset, the highest performance is \textbf{bold}, and the second highest is \underline{underlined}.}
\label{table:model_performance}
\vspace{-0.4cm}
\end{table*}

\section{Results} \label{sec: results}
In this section, we evaluate TBMs from four aspects: (1) Sec \ref{sec: end-to-end} validates the end-to-end performance of TBMs against black-box baselines; (2) Sec \ref{sec: Concept Generation Module Evaluation} assess the quality of generated concepts through human evaluation; (3) Sec \ref{sec: Concept Measurement Module Evaluation} calculates the correlation between human and TBMs on concept measurement and (4) Sec \ref{sec: analysis} demonstrates how the interpretability of TBMs provide a novel way to analyze the learning of the model.
\subsection{End-to-End Performance} \label{sec: end-to-end}
% \begin{table*}[!t]
% \centering
% \resizebox{\textwidth}{!}{%
% \begin{tabular}{l|cccc|cccc}
% \toprule
% \multirow{2}{*}{\begin{tabular}[c]{@{}c@{}}\textbf{Model}  (Interpretable)\end{tabular}} & \multicolumn{4}{c|}{\textbf{Natural Language Inference}} & \multicolumn{4}{c}{\textbf{Topic/Miscellaneous}} \\ \cmidrule{2-9}

% & \textbf{SNLI} & \textbf{MNLI} & \textbf{QNLI} & \textbf{Avg} &  \textbf{AG} & \textbf{SciCite}  & \textbf{Phishing} & \textbf{Avg} \\ \midrule

%  Naive Bayes (\checkmark) & 0.361 & 0.320 & 0.540 & 0.407 & 0.852 & 0.565 & 0.873 & 0.763 \\  \midrule
% BERT-base (\xmark) & 0.716 & 0.628 & 0.769 & 0.704 & 0.907 & 0.860 & 0.911 & 0.893 \\
% DeBERTa-base (\xmark) & 0.795 & 0.751 & 0.804 & 0.783 & 0.906 & 0.875 & 0.928 & 0.903 \\
% GPT-3.5 (5-shot) (\xmark) & 0.514 & 0.476 & 0.514 & 0.501 & 0.834 & 0.494 & 0.915 & 0.748 \\
% GPT-3.5 (fine-tune) (\xmark) & 0.869 & 0.800 & 0.899 & 0.856 & 0.873 & 0.839 & 0.937 & 0.883 \\  \midrule

% TBM (Ours) (\checkmark) & 0.593 & 0.514 & 0.641 & 0.583 & 0.796 & 0.590 & 0.889 & 0.758  \\
% $\Delta$ (\%) vs. few-shot & +15.37 & +7.98 & +24.71 & +16.22 & -4.56 & +19.43 & -2.84 & +1.43\\
% $\Delta$ (\%) vs. fine-tune & -31.76 & -35.75 & -28.70 & -31.93 & -12.24 & -32.57 & -5.12 & -16.33\\
% \bottomrule
% \end{tabular}
% }
% \caption{Model performance on 3 \textbf{NLI} datasets and 3 \textbf{Topic/Miscellaneous} datasets. We report test accuracy for all datasets.}
% \label{table:model_performance_misc}
% \vspace{-0.3cm}
% \end{table*}

\begin{table*}[!t]
\centering
\small
\begin{tabular}{l|ccccccc}
\toprule
\textbf{Model} (Interpretable) & \textbf{SNLI} & \textbf{AG} & \textbf{Patent} & \textbf{SciCite} & \textbf{Fake} & \textbf{Partisan} & \textbf{Avg} \\ \midrule
BERT-base (\xmark) &  0.480 & \textit{0.904} & 0.462 & 0.692 & 0.705 & 0.784 & 0.671\\
DeBERTa-base (\xmark) & 0.512 & \textbf{0.912} & \underline{0.488} & \textbf{0.776} & \underline{0.846} & 0.776 & 0.718\\
GPT-3.5 (fine-tune) (\xmark) & \underline{0.864} & \textit{0.904} & \textbf{0.604} & \underline{0.764} & \textbf{0.950} & \textbf{0.852} & \textbf{0.823} \\
GPT-4 (10-shot) (\xmark) & \textbf{0.868} & \underline{0.908} & \textit{0.474} & 0.688 & 0.842 & \underline{0.808} & \underline{0.765} \\ \midrule
 Naive Bayes (\checkmark) & 0.368 & 0.716 & 0.282 & 0.512 & 0.419 & 0.668 & 0.494 \\
TBM (Ours) (\checkmark) &  \underline{0.864} & 0.860 & 0.464 & \textit{0.724} & \textit{0.755} & \textit{0.804} & \textit{0.745}\\
\bottomrule
\end{tabular}
\caption{Model accuracy on \textbf{non-sentiment} domains: natural language inference (SNLI), topic classification (AG, Patent), intent detection (SciCite), and news analysis (Fake, Partisan). The best performance is \textbf{bold}, the second best is \underline{underlined}, and the third best is \textit{italic}.}
\label{table:model_performance_misc}
\vspace{-0.4cm}
\end{table*}

\textbf{TBMs perform competitively with black-box baselines on sentiment domains.}
% As shown in Table~\ref{table:model_performance}, TBMs achieve the second-highest average accuracy across all sentiment classification datasets (0.89) and the second-lowest average MSE (0.713) across all regression datasets, surpassing all the baselines except for finetuned GPT-3.5. Compared to black-box baselines such as few-shot GPT-4 and finetuned DeBERTa, TBMs exhibit competitive and consistent performance. These results are particularly surprising given that, compared to black-box models, TBMs have access to much less information due to the concept bottleneck while still maintaining the performance. By contrast, the interpretable baseline, Naive Bayes, falls far behind. 
Table~\ref{table:model_performance} shows the performance of TBMs on various sentiment-related tasks. TBMs achieve the highest or second-highest performance on 4 datasets. Looking at the gap when TBMs are outperformed by the black-box baseline, we see that on regression datasets, the MSE of the best-performing datasets, on average, is 8.7\% higher than the TBM MSE. On classification datasets where the TBMs underperform the best model, the average performance difference is 1.4\%. Overall, this indicates minimal interpretability-performance trade-offs for the sentiment domain.

\noindent \textbf{TBMs struggle on non-sentiment domains.}
We note that the structure of the TBM can naturally perform well for sentiment-related tasks since it is easier to decompose sentiment measurement into independent subtasks. Concepts for sentiment can be discovered in any order, and the measurement of one concept is not typically expected to impact the measurement of another concept. Some domains, such as fake news detection, where one concept measurement (e.g, the factuality of a claim) might change depending on another concept (the time that the claim was made), might not as be easily modeled by the current bottleneck structure due to the lack of concept hierarchy. 
% To examine how well the model can generalize across different NLP tasks, we evaluate its performance on diverse domains, including natural language inference (SNLI), topic classification (AG News, Patent), intent detection (SciCite), and news analysis (Fake News, News Partisanship). 
Table~\ref{table:model_performance_misc} shows the TBM performance across diverse domains. In contrast to the sentiment domain, TBMs underperform across all tasks except for SNLI, where the gap between the TBM and the best model is only 0.4\%. TBMs are modestly outperformed, near 5\% of the best model, in AG News (5.2\% gap), Partisan News (4.8\%), and SciCite (5.2\%). Finally, they are significantly outperformed for Patent Classification (14\% gap) and Fake news detection (19.5\% gap). These results potentially indicate that some tasks may be difficult for TBMs to decompose into linearly interacting concepts.

\subsection{Human Evaluation on Concept Generation Module}
\label{sec: Concept Generation Module Evaluation}

\begin{figure*}[!b]
\vspace{-0.4cm}
\begin{center}

\includegraphics[width=\textwidth]{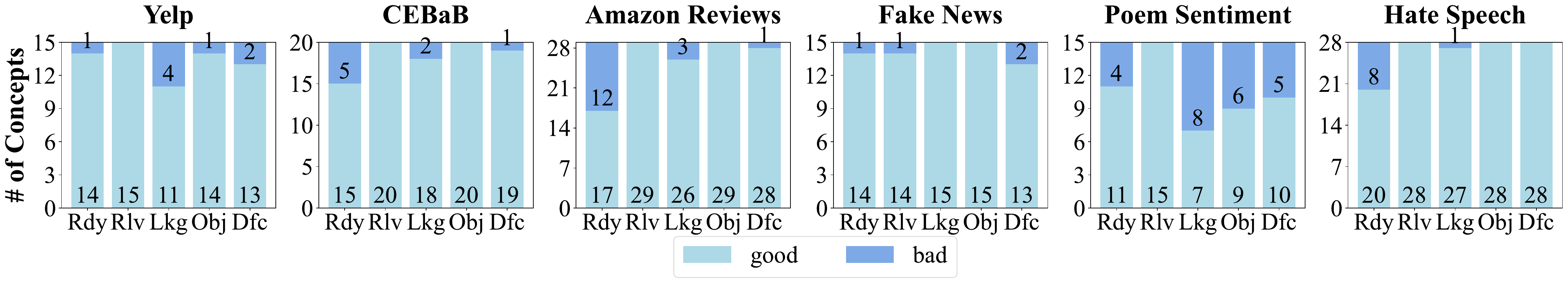}
\end{center}
\vspace{-.4cm}
\caption{Expert concept annotations for concept generation quality on five aspects: 
\textbf{Redundancy (Rdy)} is concept duplication,  ``bad'' indicates repetition; 
\textbf{Relevance (Rlv)} is pertinence to the task, ``bad'' identifies spurious concepts; 
\textbf{Leakage (Lkg)} checks if the concept directly performs the task, ``bad'' indicates leakage; 
\textbf{Objectivity (Obj)} is measurability clarity, with ``bad'' indicates subjectivity; and 
\textbf{Difficulty (Dfc)} checks the complexity of measuring the concept, ``bad'' means the concept measurement is harder than dataset task.}
\label{fig:expert_concept_annotations}
\vspace{-.3cm}
\end{figure*}

\textbf{Setup.} To assess the Concept Generation module, we manually evaluate the generated concepts along 5 aspects: \textit{Redundancy}, \textit{Relevance}, \textit{Leakage}, \textit{Objectivity}, and \textit{Difficulty}, each explained in the caption of Figure~\ref{fig:expert_concept_annotations}.
Three annotators, all authors of this paper, perform this evaluation for each concept on six datasets. A simple majority vote resolves conflicts.

% \begin{wrapfigure}{r}{0.5\textwidth}

\begin{figure}[!t]
\begin{center}
\includegraphics[width=6.9cm]{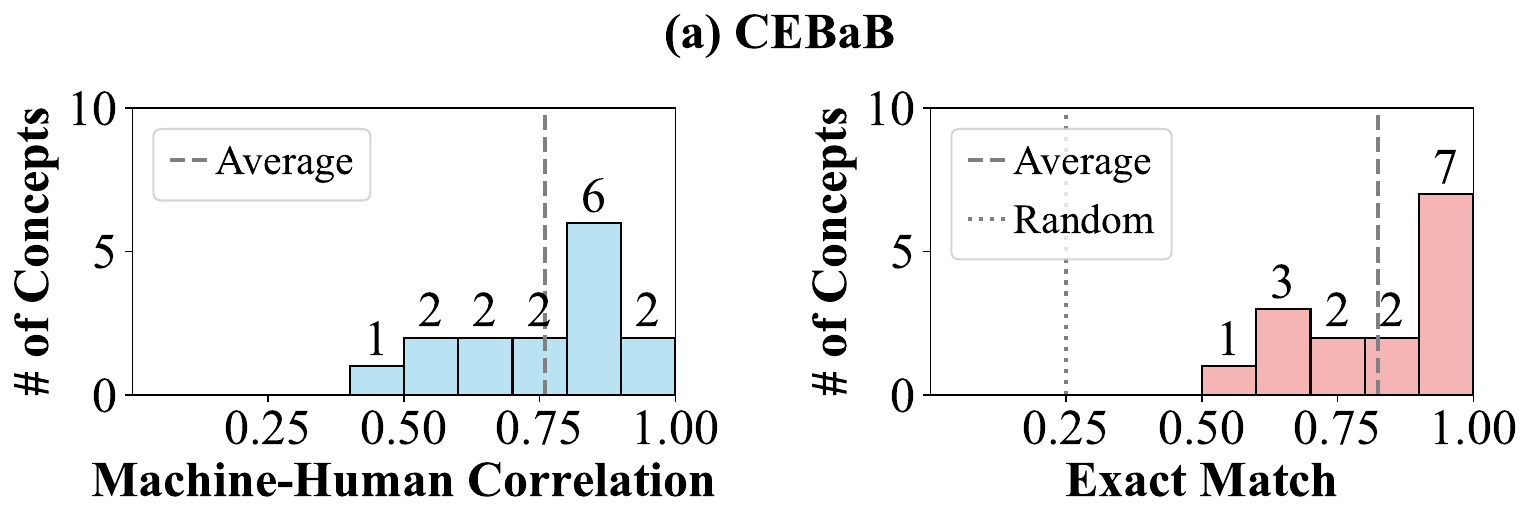}
\includegraphics[width=6.9cm]{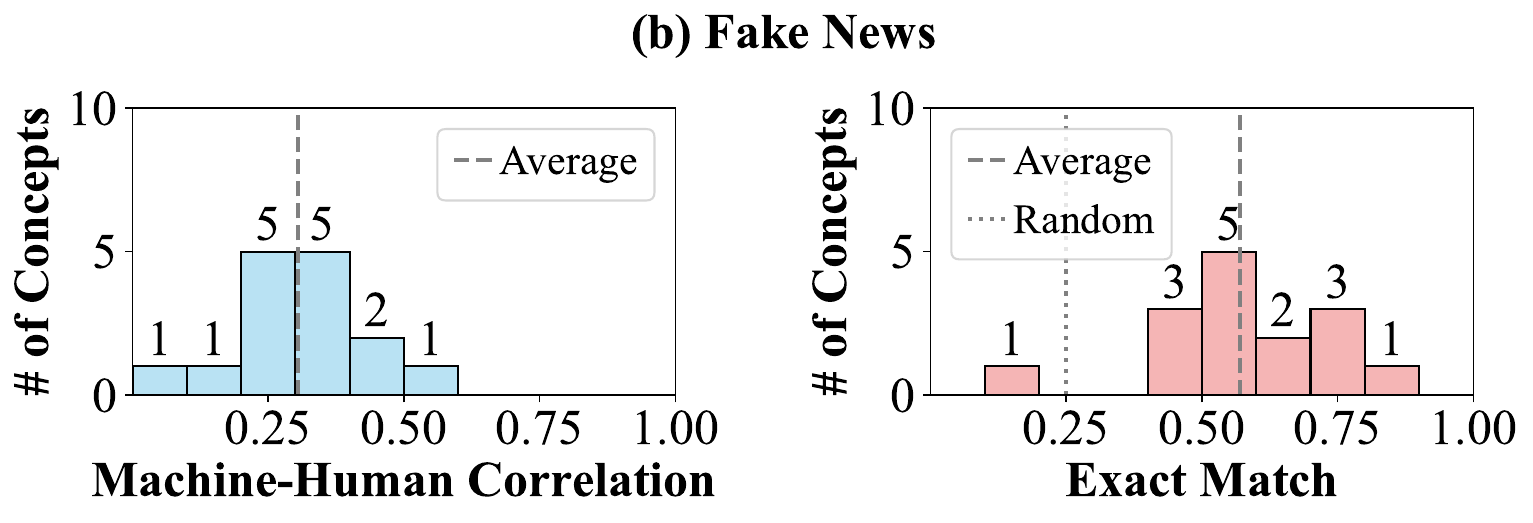}
\end{center}
\vspace{-.4cm}
\caption{Human evaluation on concept measurement. \textbf{Machine-human correlation} measures the Pearson correlation between the concept scores measured by the LLM vs. human annotators. \textbf{Exact Match} refers to the performance of the LLM in predicting the exact string label for a concept when using human annotation as ground truth.}
\label{fig:concept_measurement}
\vspace{-.4cm}
\end{figure}

% \end{wrapfigure}

\textbf{Results.} According to Figure~\ref{fig:expert_concept_annotations}, across all datasets, the overwhelming majority of concepts are of high quality, except Poem Sentiment. On average, Redundancy emerges as the most common issue (25\%), followed by Leakage (15\%), difficulty (9\%), objectivity (6\%), and relevance (1\%). This suggests that this module has almost no problem discovering concepts relevant to the task label and that most concepts are unambiguous and easy to measure. However, it occasionally accepts unnecessary concepts that are too similar to previously generated ones or concepts that directly leak the task label. The prevalence of these issues varies across datasets. For instance, Poem Sentiment shows high concept error rates in almost all aspects except relevance, while Hate Speech concepts have mostly redundancy issues. Redundant concepts unnecessarily increase the size of the concept space, which can increase the cognitive load of users trying to interpret the model behavior. Leaky concepts can undermine the faithfulness of provided explanations. This indicates a need for additional filters for flawed concepts.

\subsection{Human Evaluation on Concept Measurement Module} 
\label{sec: Concept Measurement Module Evaluation}
\textbf{Setup.} To determine whether the Concept Measurement Module measures concepts correctly, we compare the concept scores rated by the LLM with those rated by humans on the CEBaB and Fake News datasets. We asked a group of crowd workers (more details in Sec~\ref{sec: Appendix concept measurement crowd task}) to simulate the concept measurement module. We compute the exact match and correlation between the human and LLM judgments. If annotators do not have a clear majority decision for an instance, it is labeled as ``uncertain''.

\textbf{Results.} Figure~\ref{fig:concept_measurement} (a) shows the histogram of the correlations and accuracies for all the concepts in the CEBaB dataset. The median correlation and accuracy are high at 0.814 and 0.893, respectively, with the average being 0.759 for correlation and 0.824 for accuracy. This level of agreement is remarkable since concept measurement is done in a zero-shot manner, with no training data about the specific concept being measured.  In contrast, the performance for the Fake News dataset, as shown in Figure~\ref{fig:concept_measurement} (b), is modest: the median correlation and accuracy are 0.317 and 0.549, respectively, while the average scores are 0.305 for correlation and 0.571 for accuracy. Most of this reduced agreement comes from hard-to-measure concepts in the Fake News TBM, such as ``Fact Checking'', where the LLM asserts that a text can be fact-checked without access to external resources. This stark difference in performance between the two datasets reflects the transparency and auditability of TBM. The exemplary performance on the CEBaB dataset validates this module's potential and effectiveness. Simultaneously, Fake News's suboptimal results provide clear signals for potential pitfalls that require human intervention and debugging.

\subsection{Analysis of Learning Curves}
\label{sec: analysis}
One unique advantage of TBM is its interpretable structure, which allows for easier analysis. We demonstrate this by plotting the \textit{concept learning curves} of TBM on 3 datasets in Figure~\ref{fig:concept_curves}.\footnote{See Appendix~\ref{sec: Appendix all learning curves} for the learning curves on other datasets.} These learning curves show how the TBM's performance changes on the test data as it iteratively generates new concepts. The figure also shows the importance of each concept, which is calculated as the absolute value of its weight learned by the final prediction layer.\footnote{For linear regression, this is the magnitude of regression weight associated with the concept. For logistic regression, this is the average absolute value of the concept weight across all classes.} These learning curves can be contrasted with the learning curves in black-box models, where sudden increases in model performance require in-depth investigation to identify the cause of improvement. This task is easier for a TBM, e.g., discovering the ``Customer Recommendation'' concept in the Yelp dataset led to a significant drop in MSE. However, the concept of ``Mention of Overpricing'' appears less informative as the MSE remains stable after its addition.  Identifying the causes of improvement is particularly important for tasks such as identifying potential spurious cues in the model. 

\textbf{TBM for bias discovery.} In the SNLI learning curve, the model leverages concepts about only one side of the inputs, such as ``Hypothesis Truthfulness'' and ``Premise Verification'', which are well-known biases in natural language inference tasks \citep{gururangan-etal-2018-annotation, poliak-etal-2018-hypothesis}. This interesting finding suggests that the interpretable property of TBM could be beneficial in discovering biases or annotation artifacts in datasets.

In the appendix, we include two additional examples of how this added interpretability could be useful. Appendix~\ref{sec: Appendix variance analysis} shows an analysis of how the performance of different training runs on the same dataset can be explained using the discovered concepts, and Appendix~\ref{sec: Appendix small dataset training} shows how we can explain the overfitting of our TBM on a small dataset based on the discovery of spurious concepts.

\begin{figure*}[!t]
\centering
\hfill % This pushes all subsequent content to the right
\begin{subfigure}{.32\linewidth}
    \centering
    \includegraphics[height=3.65cm]{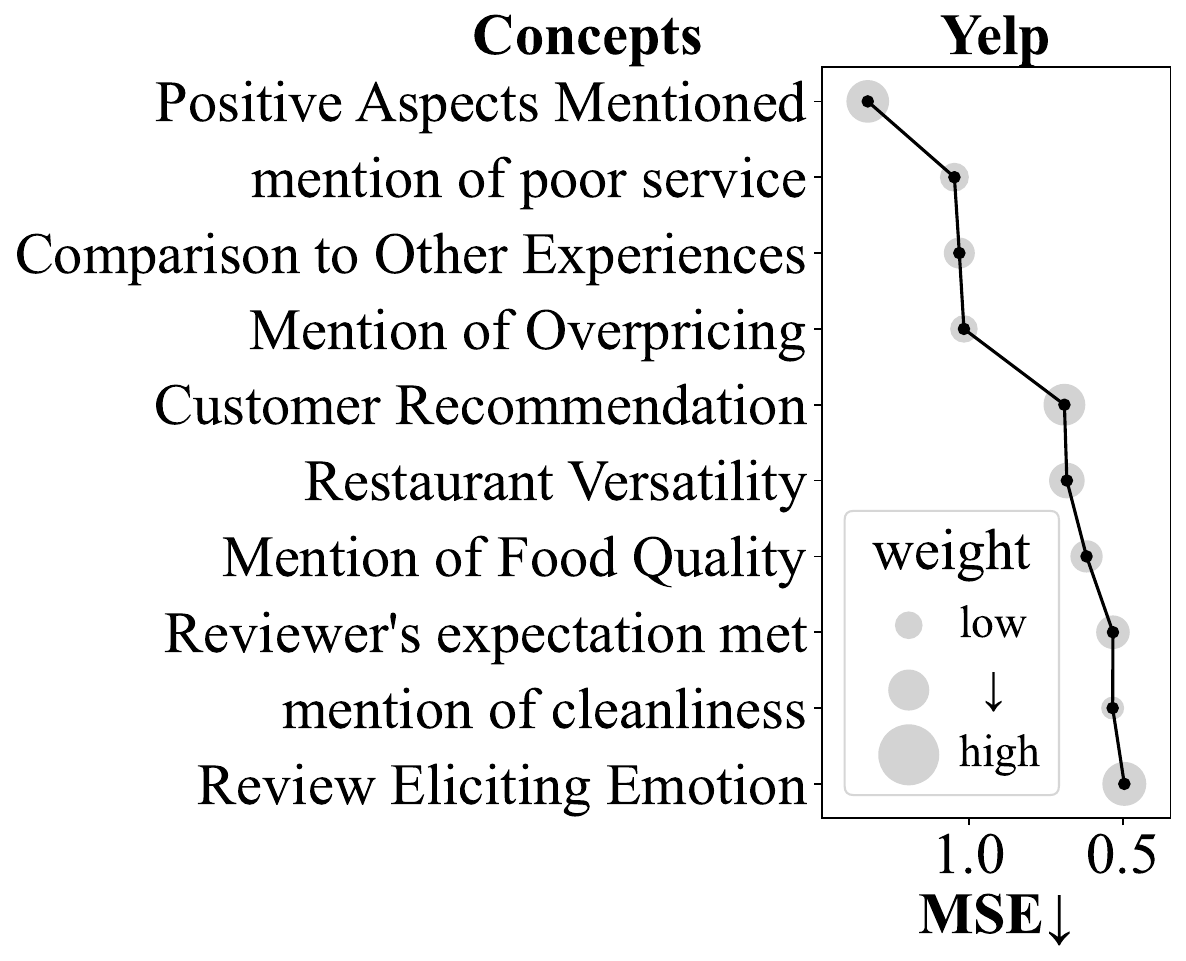}
\end{subfigure}
\hfill
\begin{subfigure}{.32\linewidth}
    \centering
    \includegraphics[height=3.65cm]{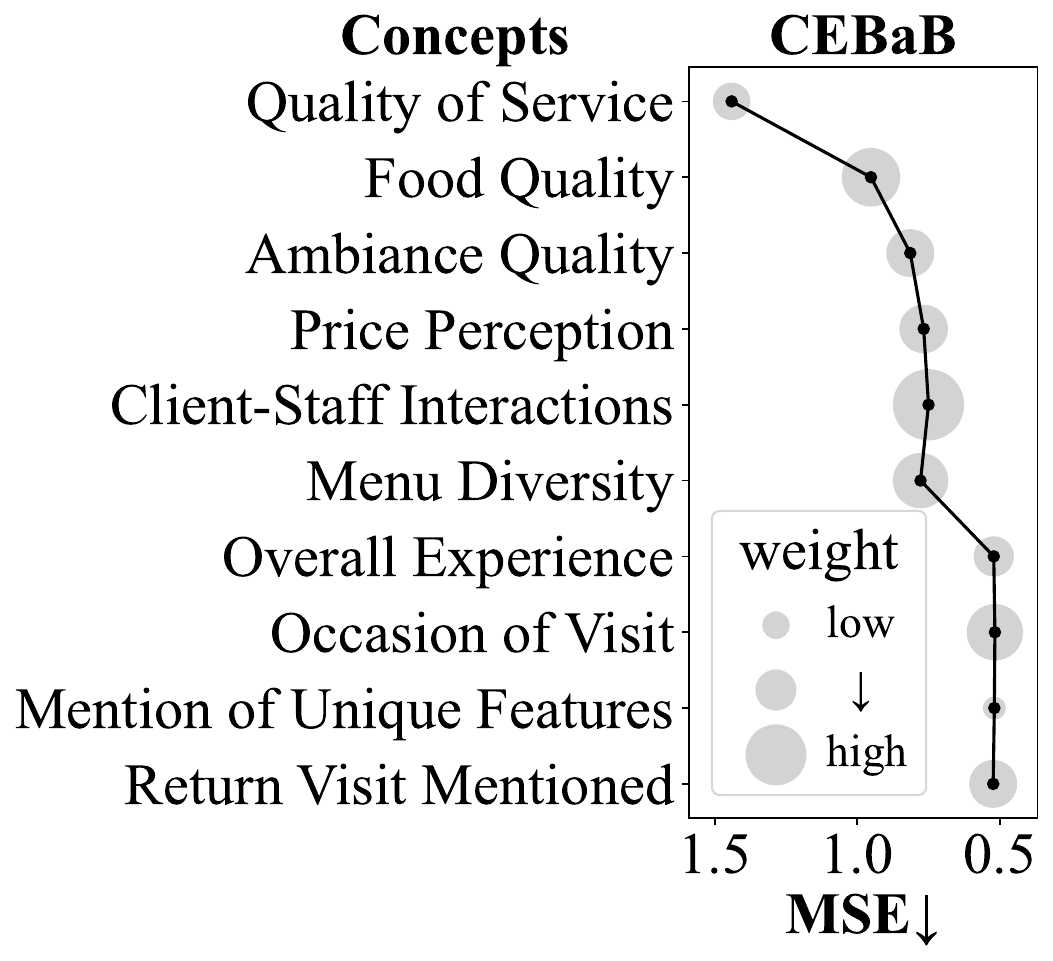}
\end{subfigure}
\begin{subfigure}{.32\linewidth}
    \centering
    \includegraphics[height=3.65cm]{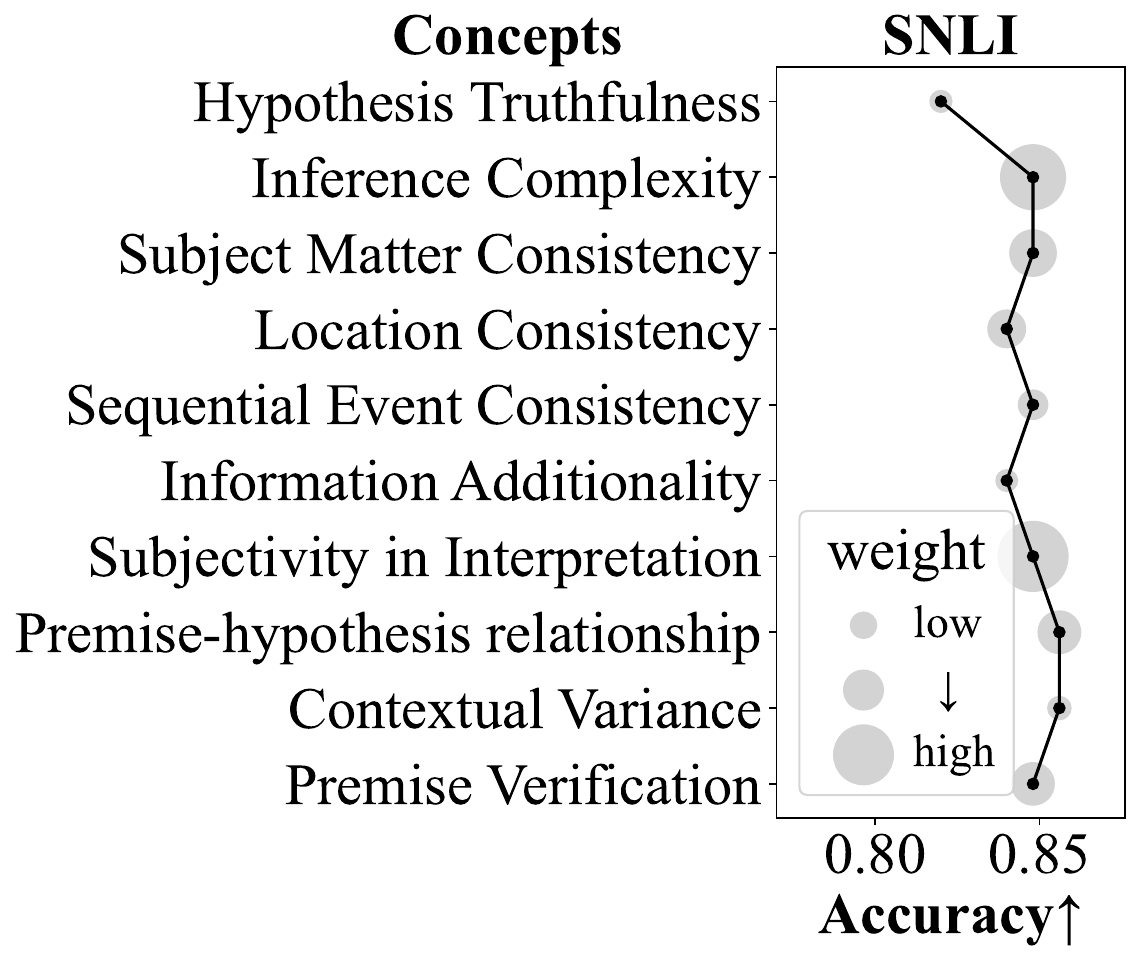}
\end{subfigure}
\hfill\null % This is a trick to prevent any issues due to the last 
\vspace{-.2cm}
\caption{Concept learning curves of TBM on 3 datasets. The x-axis represents the TBM's performance (MSE for regression task and Accuracy for classification tasks) at each iteration, and the y-axis indicates the specific concept added to the bottleneck during that iteration. The size of each node is determined by the magnitude of the weight of the corresponding concept in the prediction layer.}
\label{fig:concept_curves}
\vspace{-.4cm}
\end{figure*}

% Interestingly, we also observe that the most important concepts in terms of weight are not always discovered immediately. Instead, they can still show up at later stages of iteration. For example, in CEBaB, ``Expectations Met'' has one of the highest importance weights but is discovered last. 
\section{Conclusion and Limitations}

In this paper, we present the Text Bottleneck Model (TBM)---a fully automated interpretable-by-design text classification framework that provides both global and local interpretability with sparse concept-level explanations. In our evaluation, we show that TBM achieves competitive performance against strong black-box baselines across the sentiment domain. Human evaluations reveal that the concepts generated by the system are mostly relevant and objective, but issues of redundancy and leakage still exist. Overall, we demonstrate that TBM is a promising general architecture for constructing a highly interpretable predictor with minimal performance trade-offs.

Future work to improve the performance of the system can look into analyzing the following aspects of building TBMs:

\textbf{Scalability.}
In the current implementation for concept measurement, every time inference is performed, the number of LLM calls is proportional to the number of concepts in the bottleneck. To improve the system's scalability, it might be possible to fine-tune a smaller system for concept measurement by distilling from GPT-4 or another strong teacher model.

\textbf{Redundant and Leaky Concepts.}
The analysis of generated concepts reveals the existence of duplicate concepts and concepts that leak classification labels. To mitigate these issues, future work can include steps in concept generation to filter problematic concepts.

\noindent \textbf{Interactivity.} Since concepts are fully represented in natural language, practitioners training a TBM can easily add, delete, or edit concepts. This can allow them to steer model exploration during concept generation and refine the prompts during concept measurement. Future work can focus on how having a human in the loop can improve the system.

\bibliography{colm2024_conference}
\bibliographystyle{colm2024_conference}

\clearpage
\appendix
\appendix

\section{Further Analysis}
\subsection{Does the model generate similar concepts across repeated runs?} \label{sec: Appendix variance analysis}
\begin{figure*}
\begin{center}
\includegraphics[width=\textwidth, height=10cm]{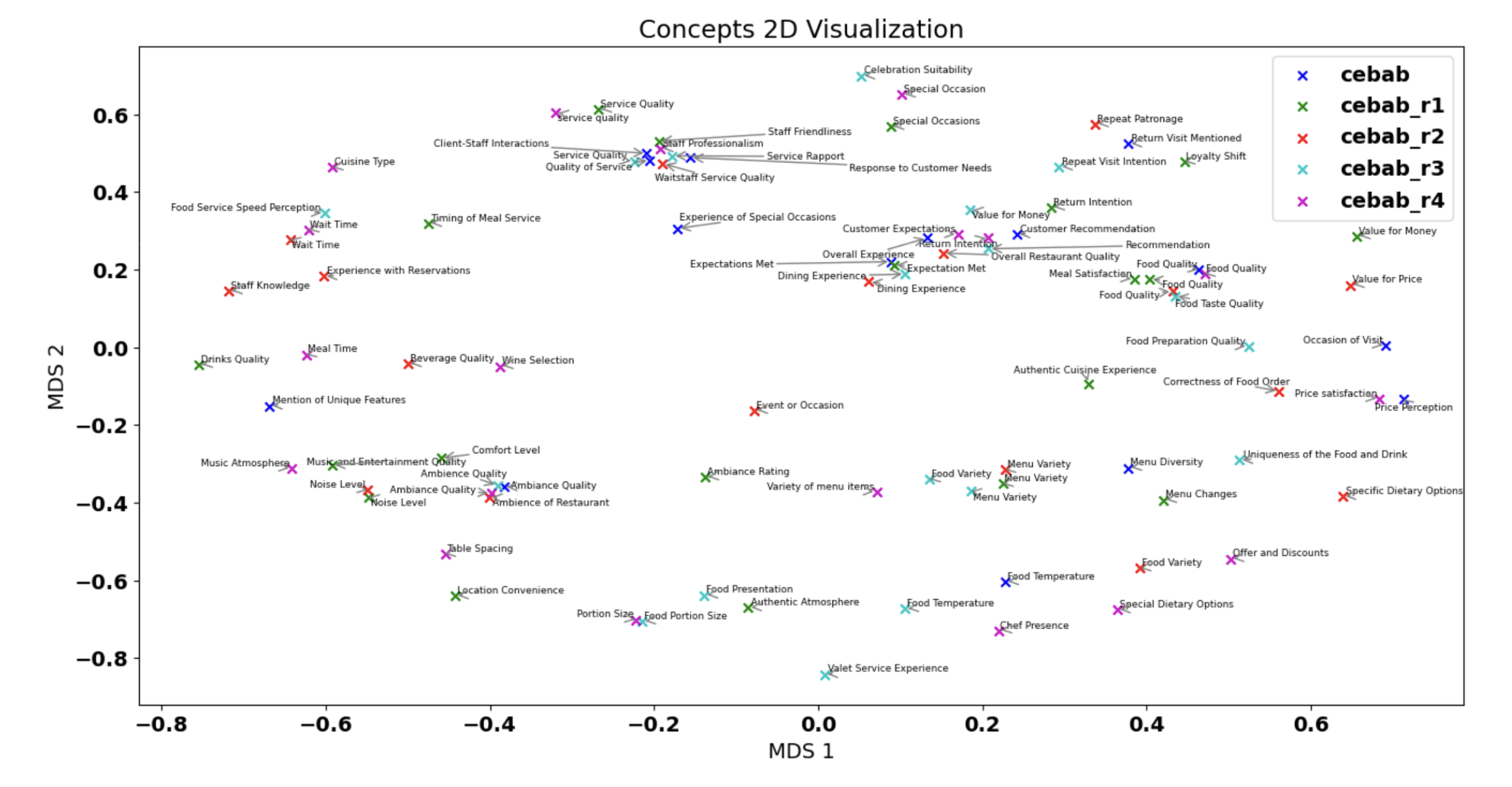}
\end{center}
\caption{Concepts across multiple CEBaB training runs, as visualized in Multi-Dimensional Scaling (MDS) Plot. The distance between concepts is proportional to their correlation. }
\label{fig:cebab_concept_MDS_plot}
\end{figure*}

\begin{figure*}
\centering
  \hfill
  \begin{subfigure}{.48\linewidth}
    \centering
    \includegraphics[height=5.5cm]{images/cebab_learning_curve.pdf}
  \end{subfigure}
\hfill
  \begin{subfigure}{.48\linewidth}
    \centering
    \includegraphics[height=5.5cm]{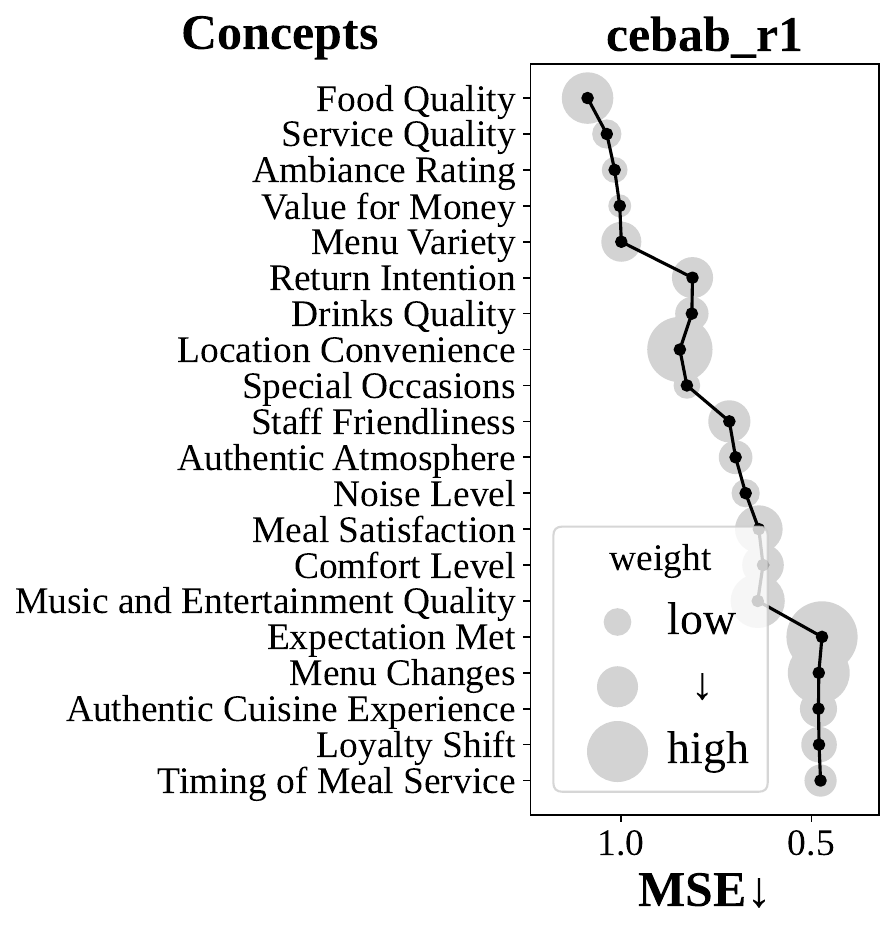}
  \end{subfigure}
  \hfill
  \begin{subfigure}{.48\linewidth}
    \centering
    \includegraphics[height=5.5cm]{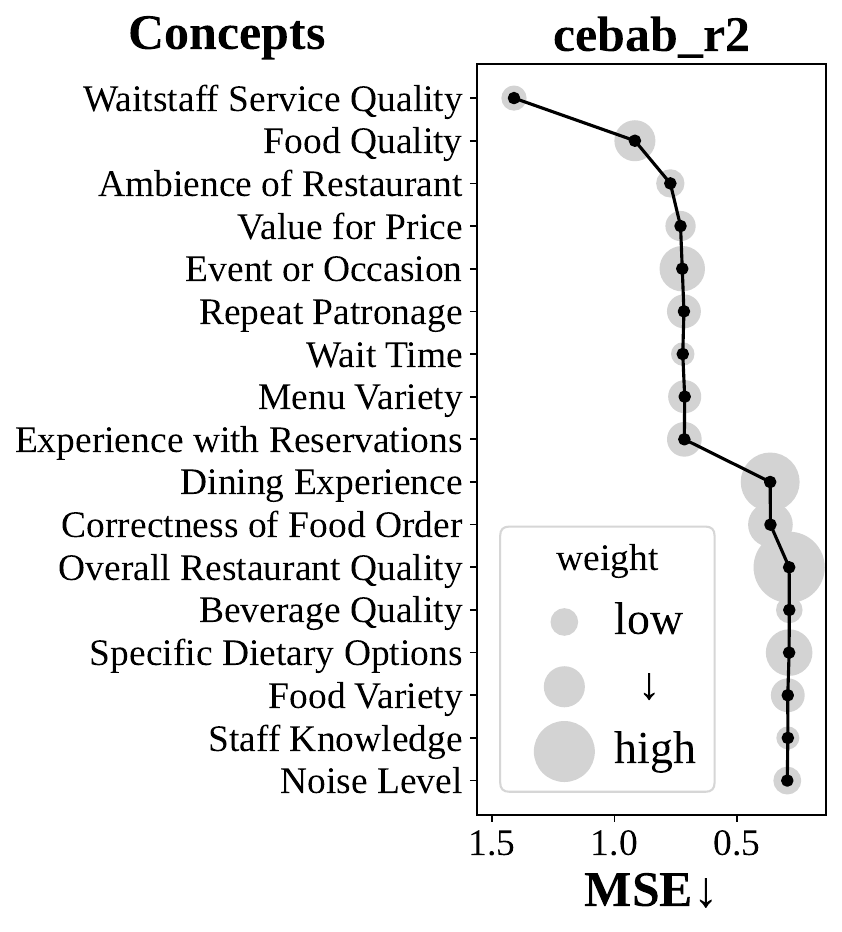}
  \end{subfigure}
  \hfill
  \begin{subfigure}{.48\linewidth}
    \centering
    \includegraphics[height=5.5cm]{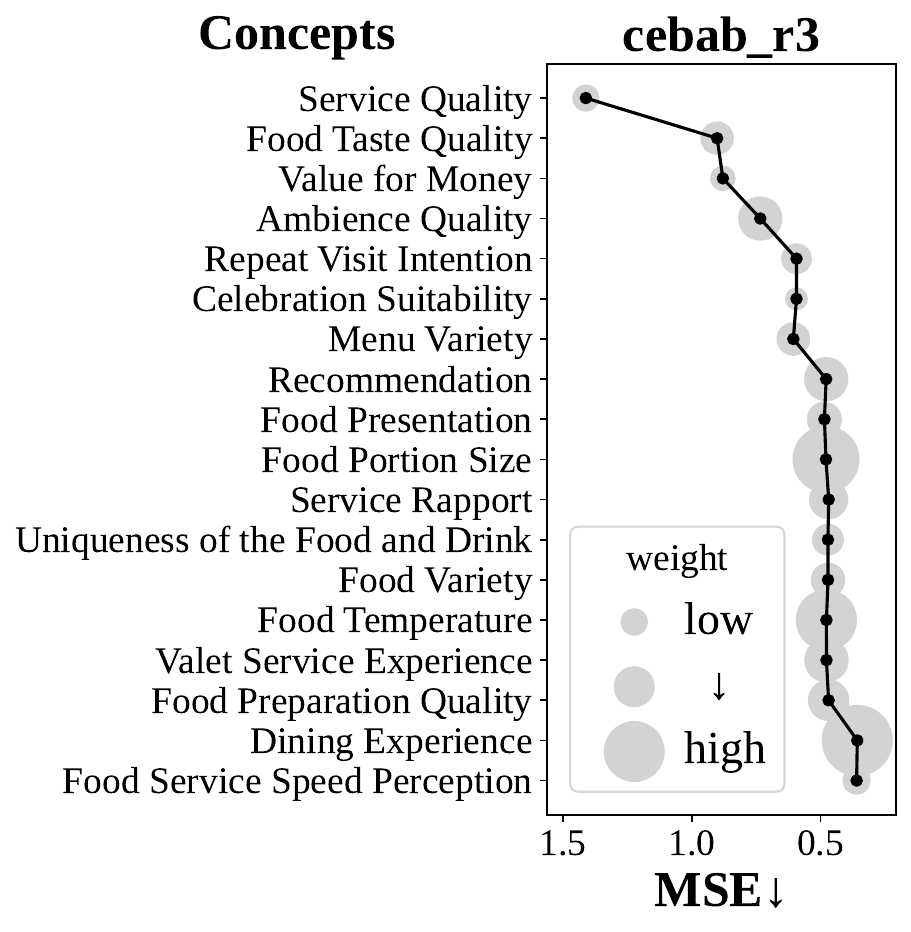}
  \end{subfigure}
  \hfill
  \begin{subfigure}{.48\linewidth}
    \centering
    \includegraphics[height=5.5cm]{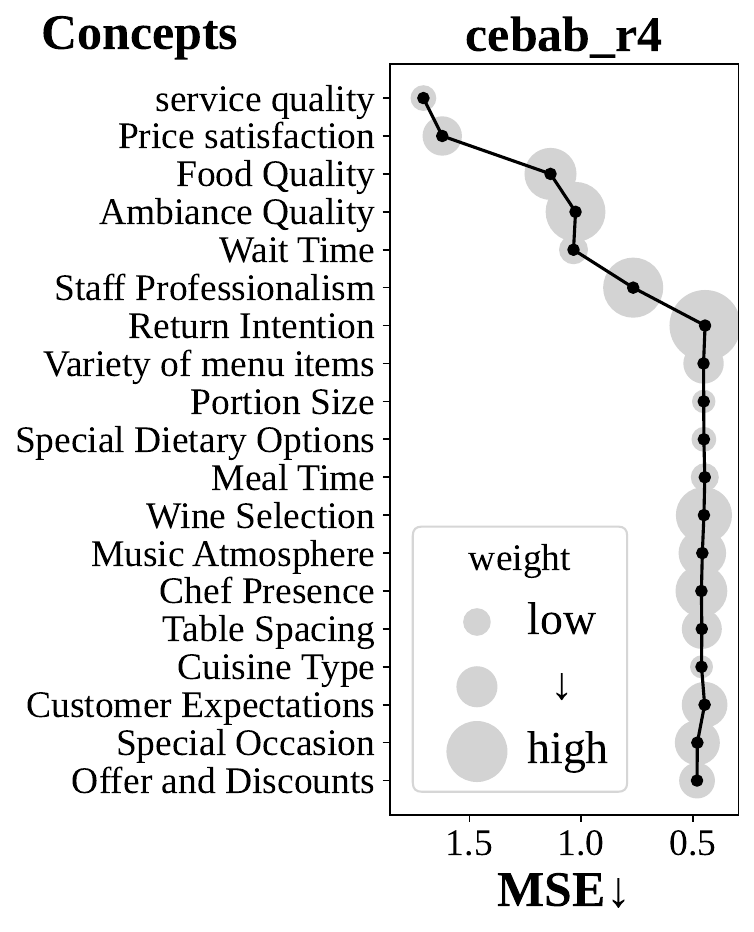}
  \end{subfigure}
    
\caption{Concepts across multiple CEBaB training runs with identical starting conditions, as visualized in learning curves.}
\label{fig:cebab_concept_curves}
\end{figure*}

To evaluate the variance in concept generation, we compare the concepts generated by a TBM on five runs on the CeBaB dataset with 250 training examples and 250 testing examples. Figure~\ref{fig:cebab_concept_MDS_plot} 
% and Figure~\ref{fig:cebab_concept_dendogram} 
visualizes the concepts generated across model runs, and Figure~\ref{fig:cebab_concept_curves} shows the learning curves for the TBMs. 
% In the dendrogram, 
We can see that concepts such as ``Menu variety'', ``Food Quality'', ``Reviewer Expectations'', ``Ambiance Quality'', ``Service Quality'' are generated by most TBMs.
Inspecting the concept learning curves reveals that these concepts are highly important in the model. Overall, these results indicate that TBMs can consistently discover the important concepts across replicated model runs.

The MSE for the final models are 0.43, 0.48, 0.29, 0.36, and 0.48, respectively. We note that both models that achieve the best performance contain concepts that leak the task label, such as ``Dining Experience'' and ``Overall Restaurant Quality''.

\begin{figure*}[!t]
\centering
  \hfill
  \begin{subfigure}{.48\linewidth}
    \centering
    \includegraphics[height=5cm]{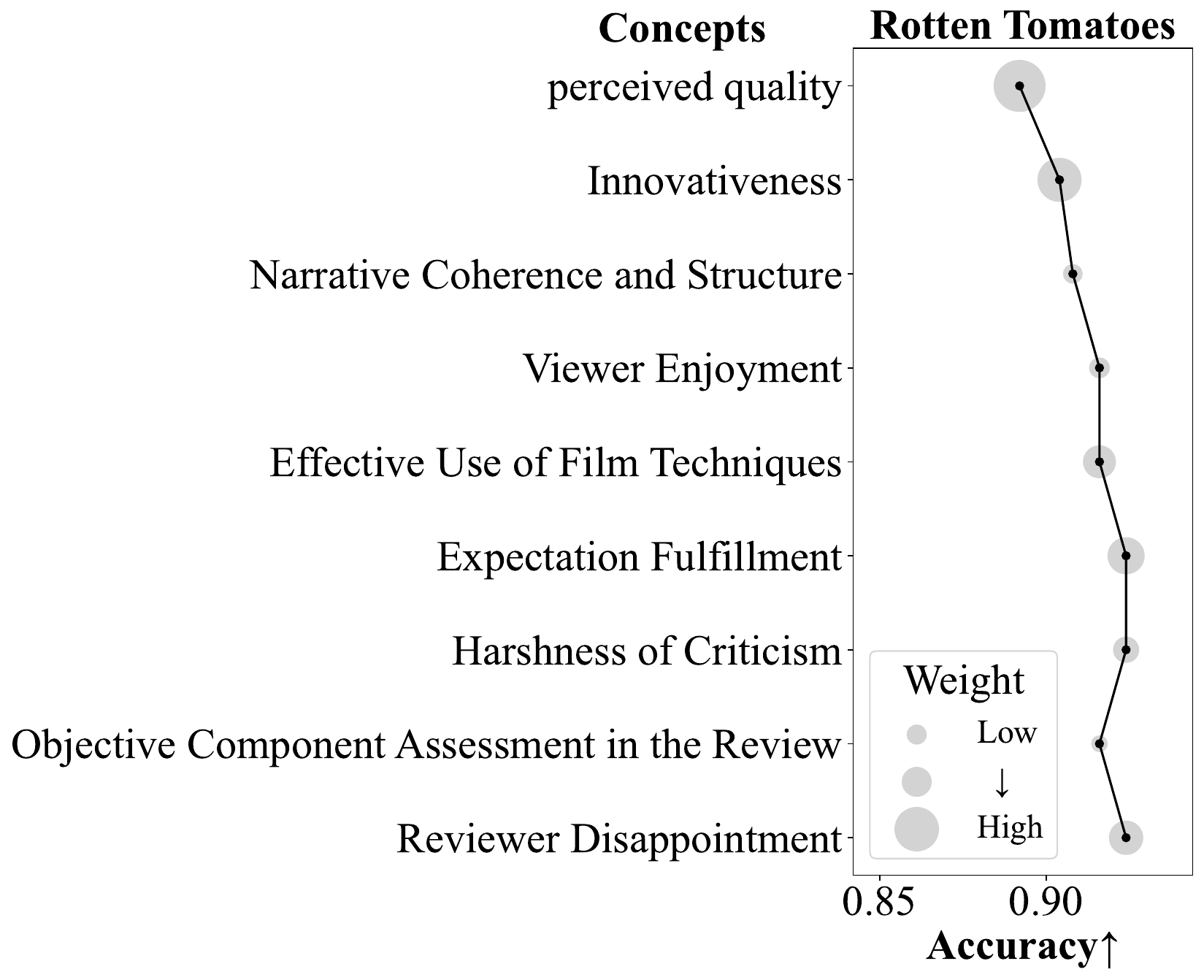}
  \end{subfigure}
\hfill
  \begin{subfigure}{.48\linewidth}
    \centering
    \includegraphics[height=5cm]{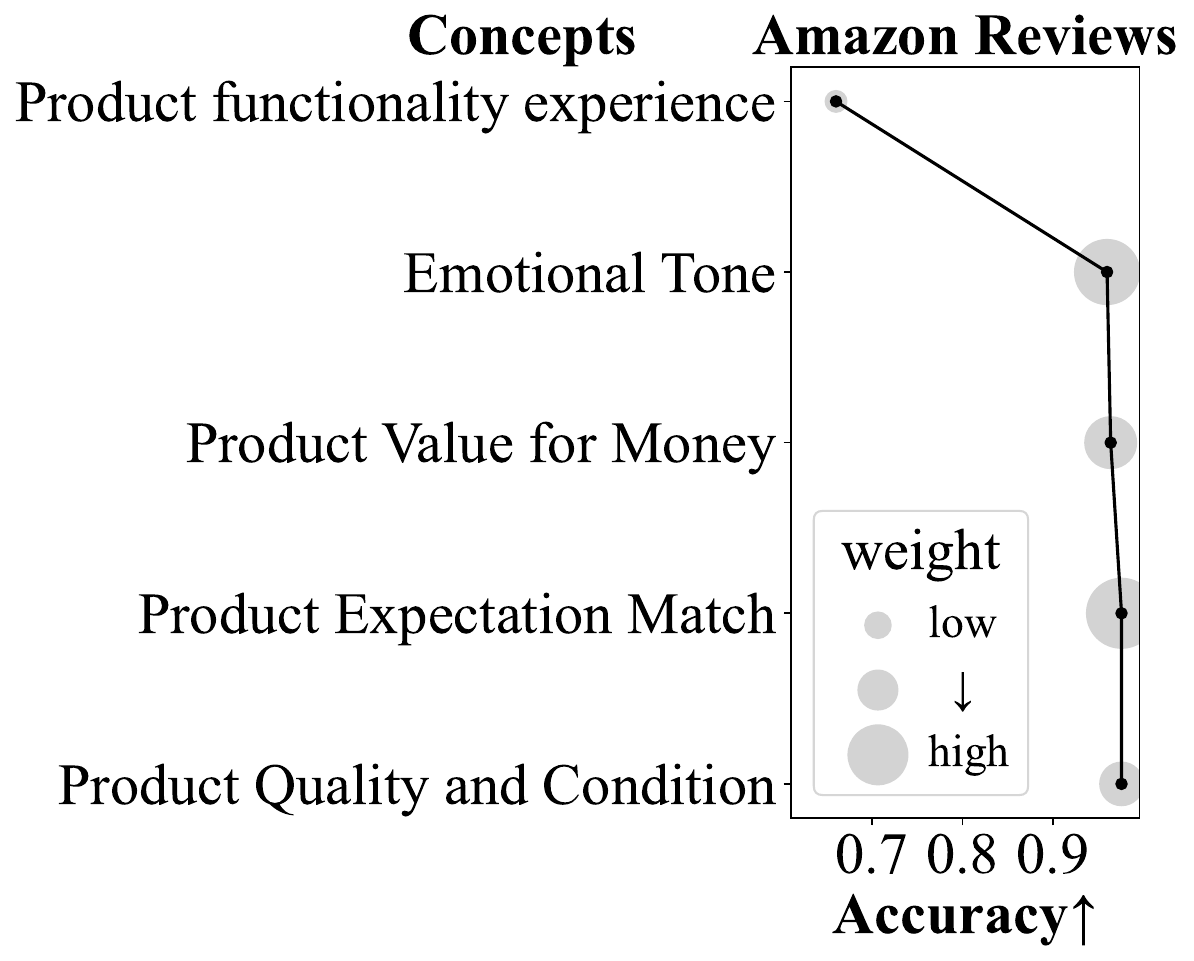}
  \end{subfigure}
  \hfill
  \begin{subfigure}{.48\linewidth}
    \centering
    \includegraphics[height=5cm]{images/Snli_learning_curve.pdf}
  \end{subfigure}
  \hfill
  \begin{subfigure}{.48\linewidth}
    \centering
    \includegraphics[height=5cm]{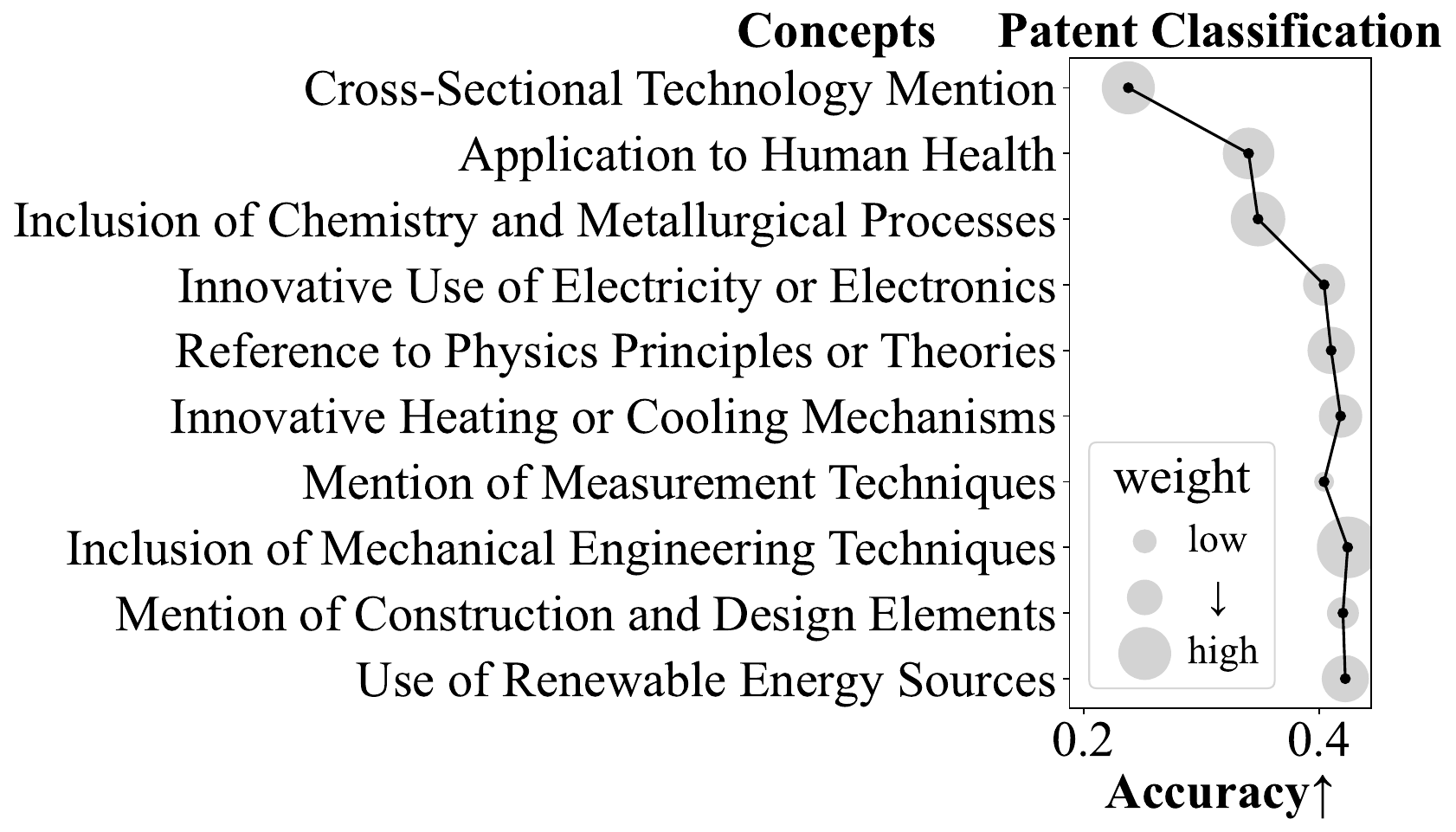}
  \end{subfigure}
  \hfill
  \begin{subfigure}{.48\linewidth}
    \centering
    \includegraphics[height=5cm]{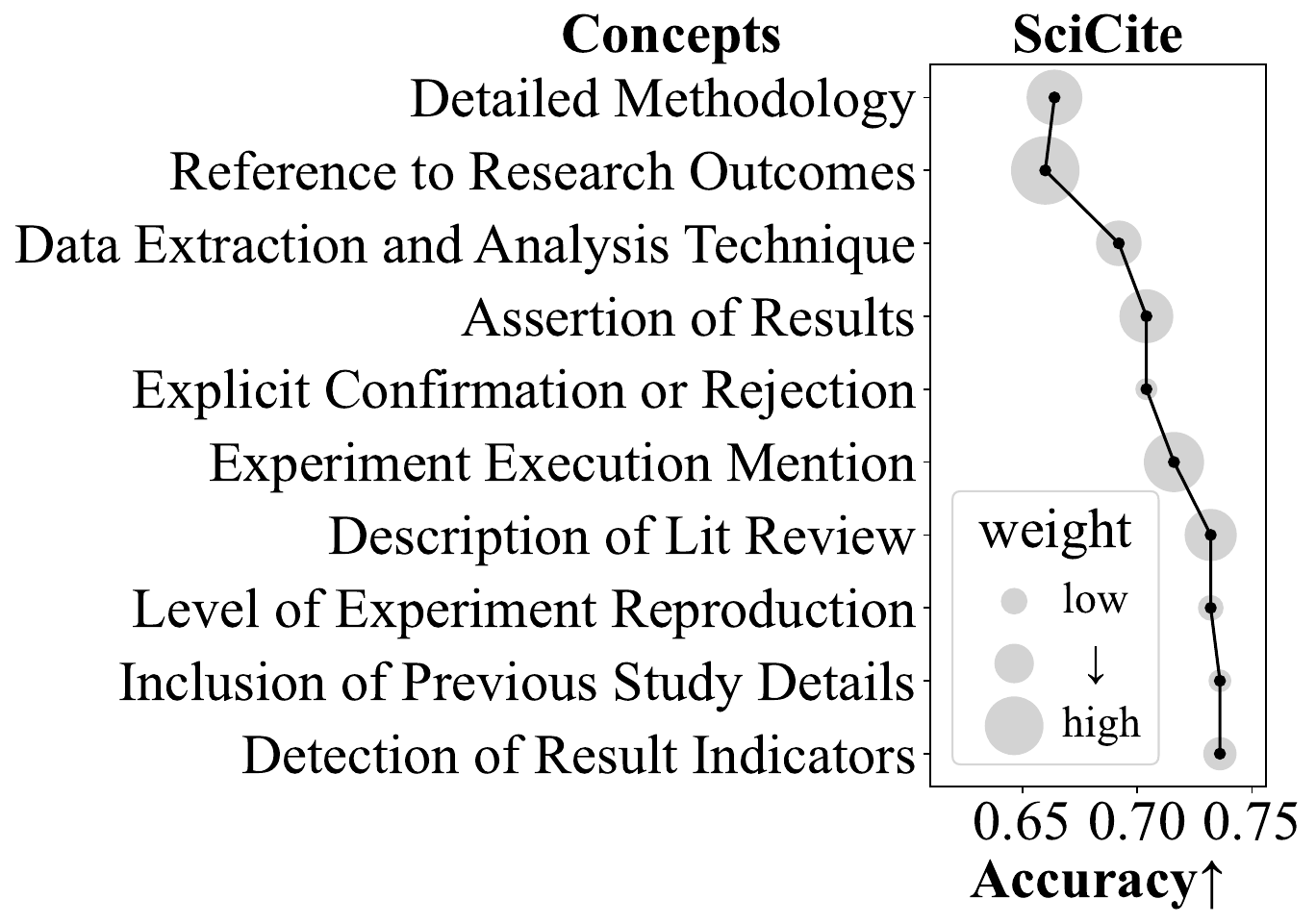}
  \end{subfigure}
    \hfill
  \begin{subfigure}{.48\linewidth}
    \centering
    \includegraphics[height=5cm]{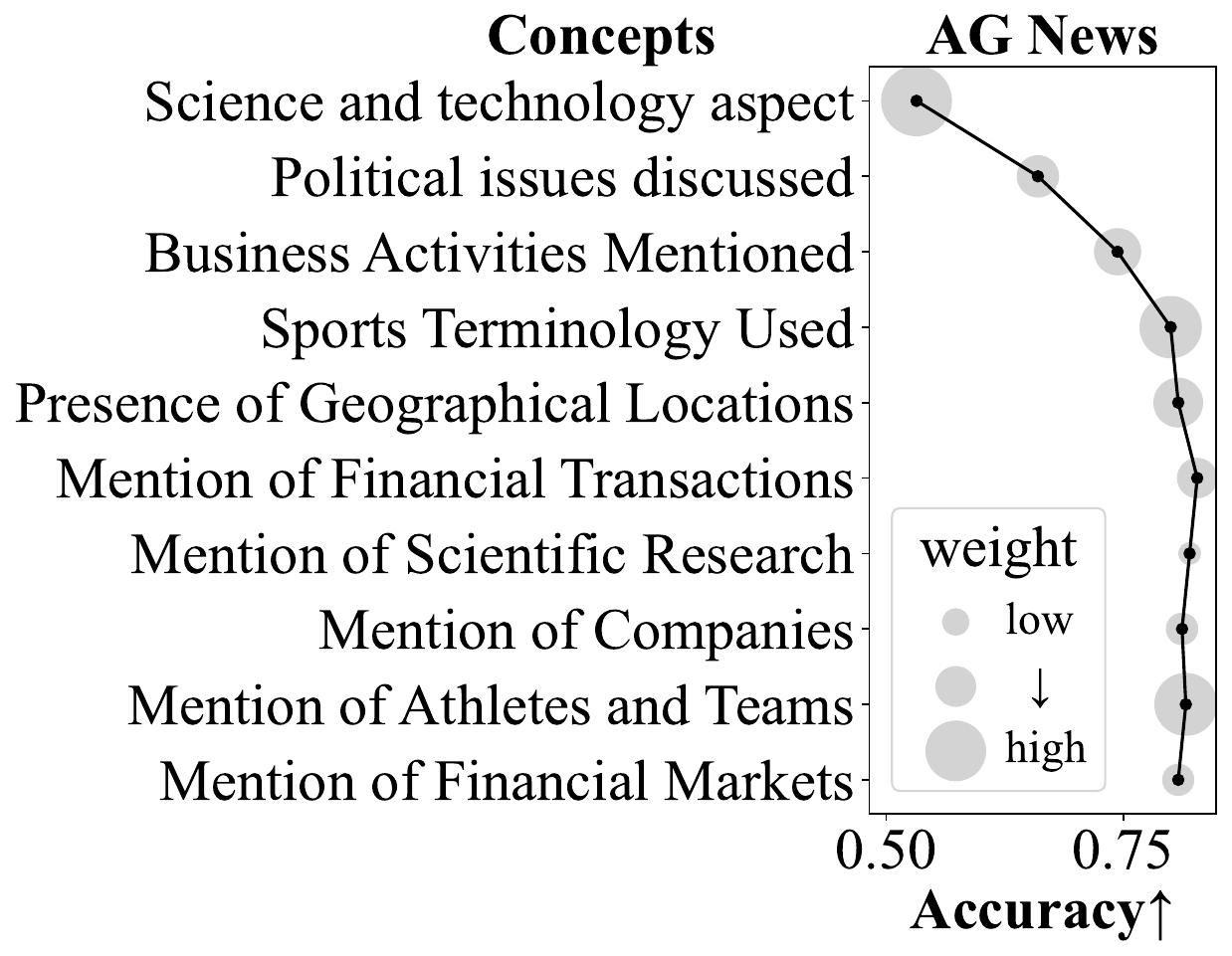}
  \end{subfigure}
    \hfill
  \begin{subfigure}{.48\linewidth}
    \centering
    \includegraphics[height=5cm]{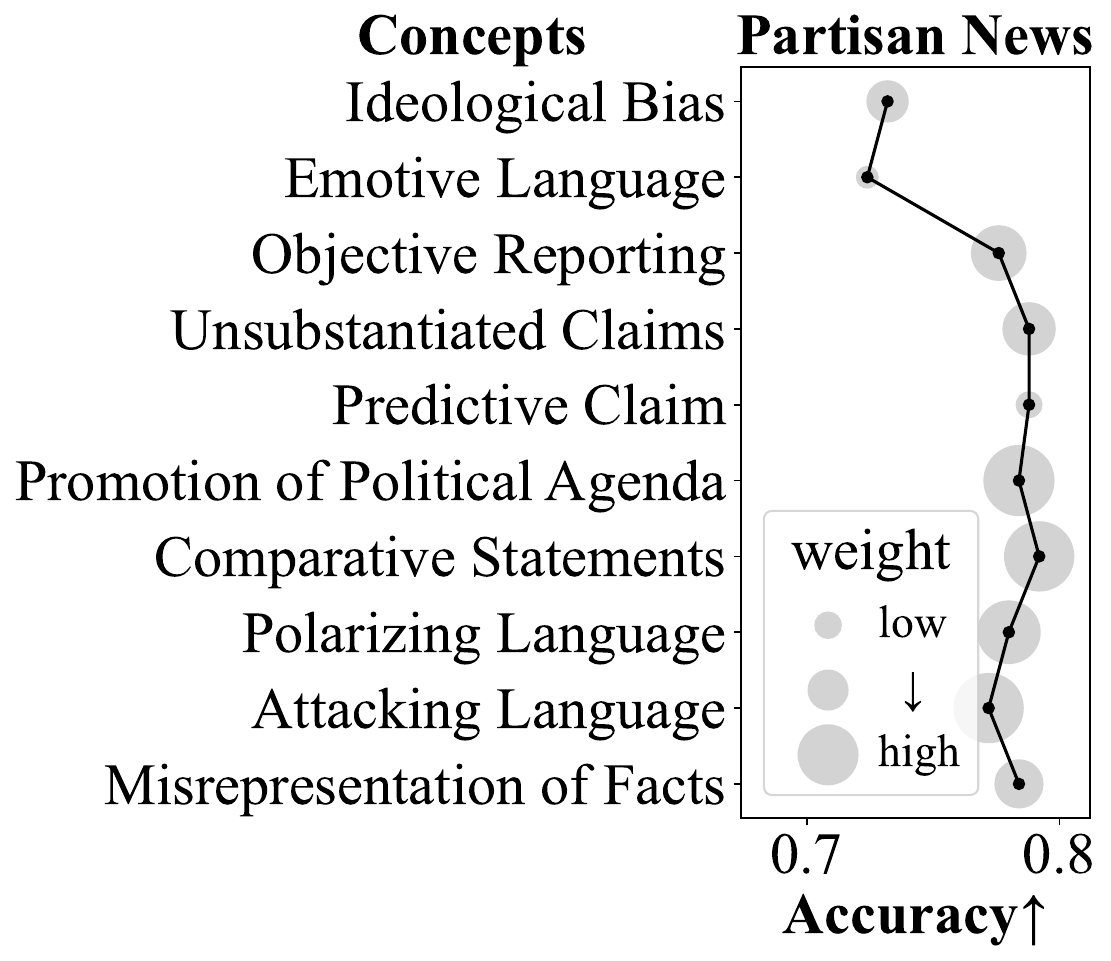}
  \end{subfigure}
      \hfill
  \begin{subfigure}{.48\linewidth}
    \centering
    \includegraphics[height=5cm]{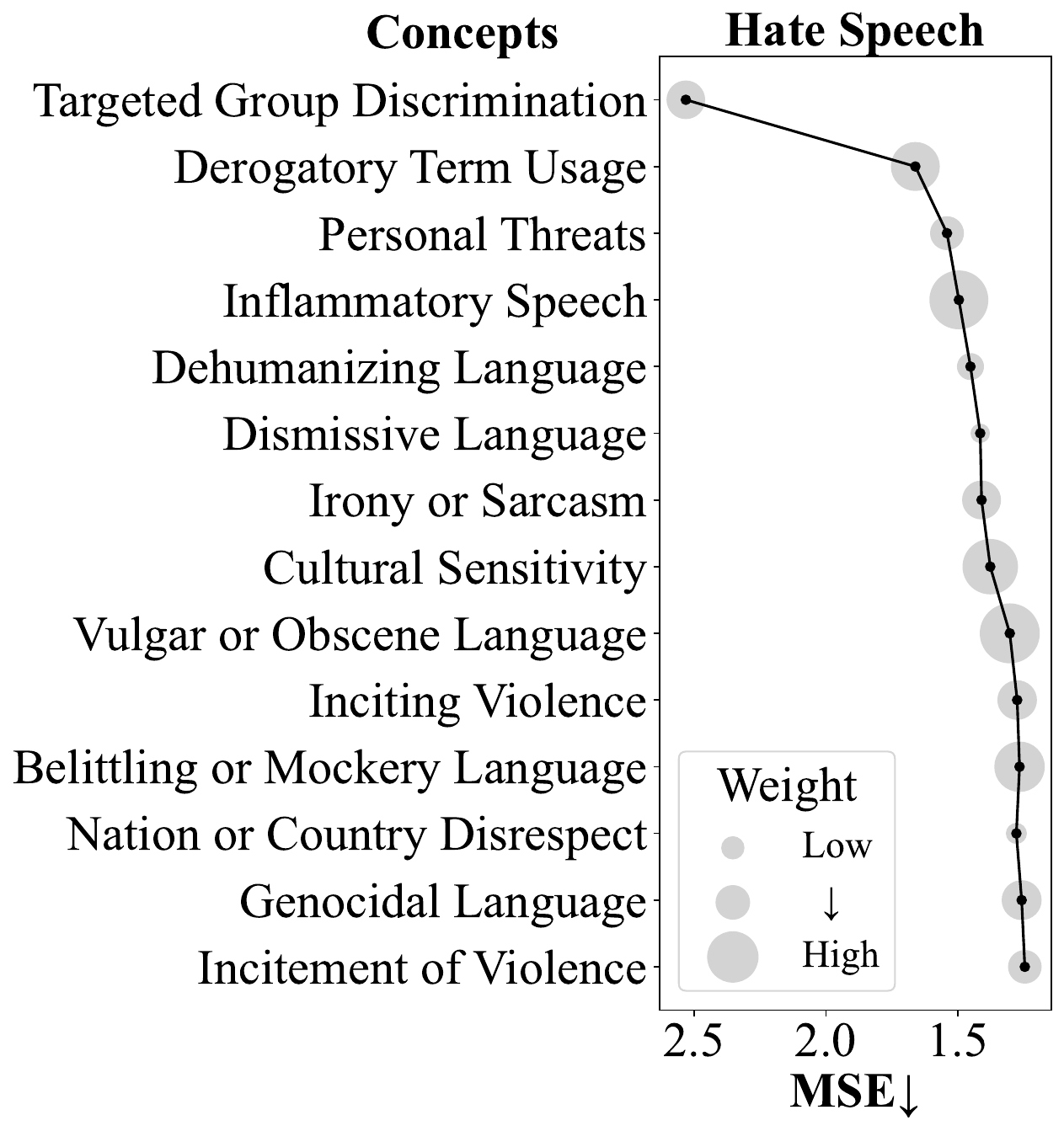}
  \end{subfigure}

\caption{Concept learning curves on all datasets.}
\label{fig:concept_curves_all}
\end{figure*}

\begin{figure*}[!t]
\centering
  \begin{subfigure}{.32\linewidth}
    \centering
    \includegraphics[height=5.0cm]{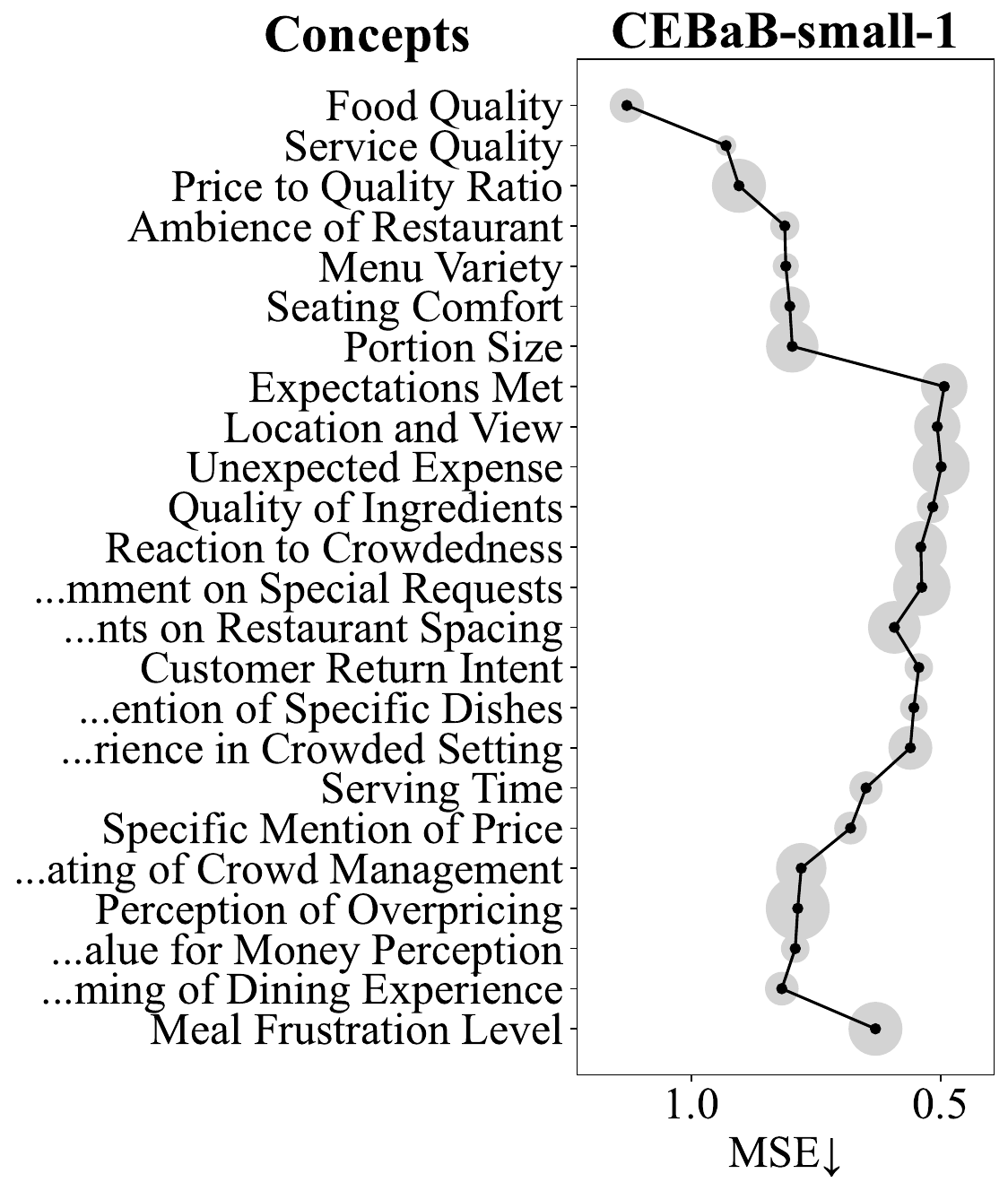}
  \end{subfigure}
  \hfill
  \begin{subfigure}{.32\linewidth}
    \centering
    \includegraphics[height=5.0cm]{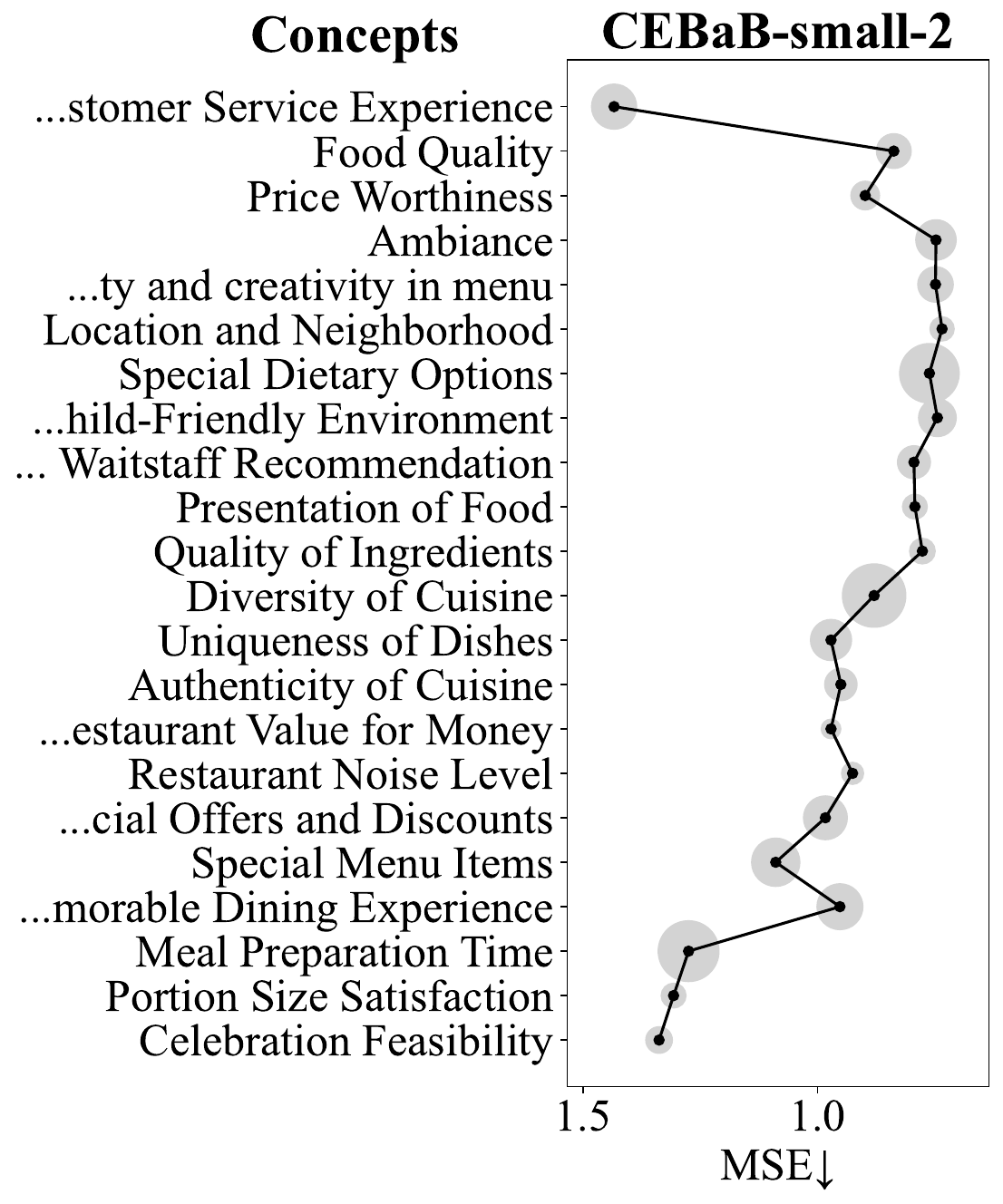}
  \end{subfigure}
  \hfill
  \begin{subfigure}{.32\linewidth}
    \centering
    \includegraphics[height=5.0cm]{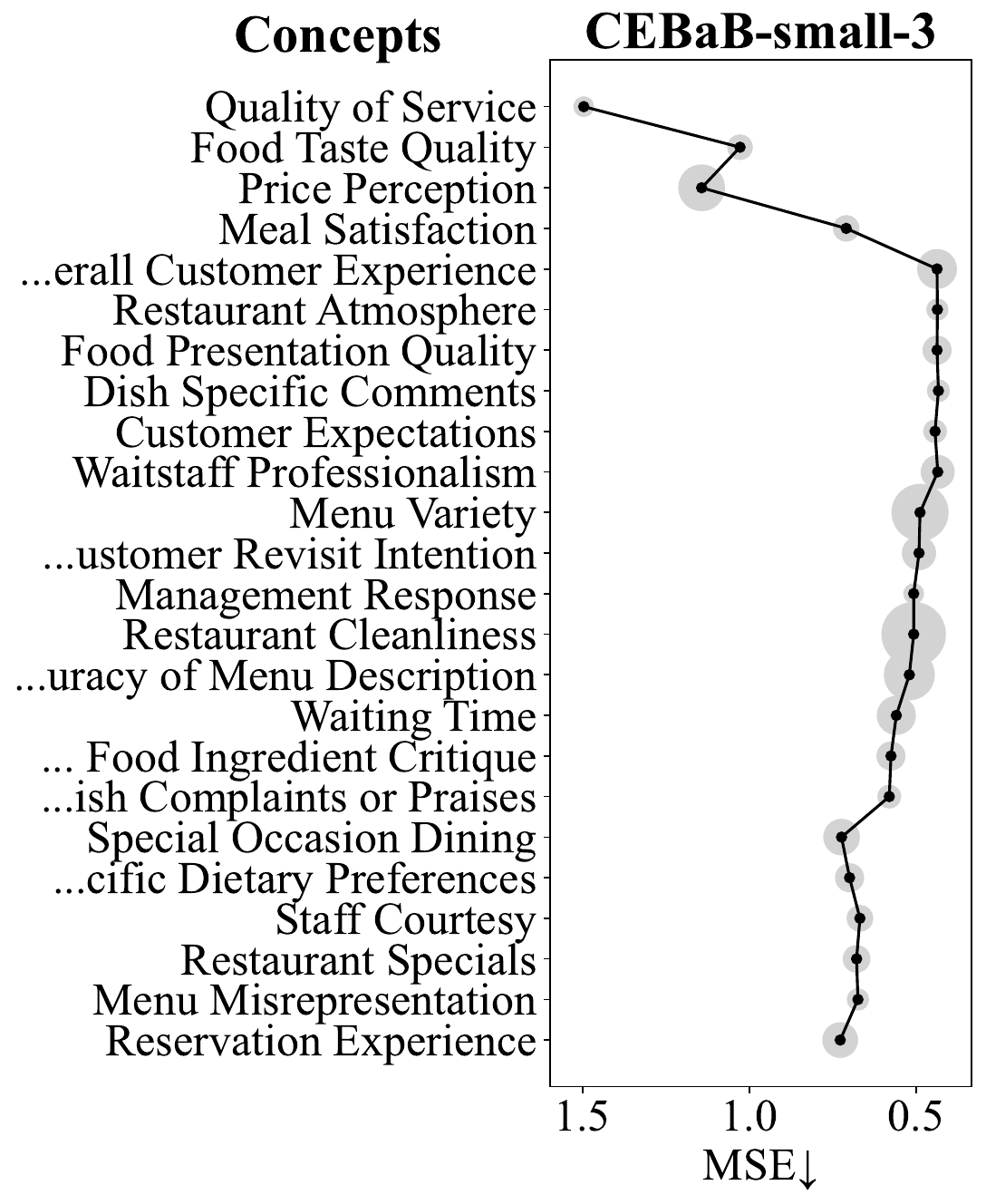}
  \end{subfigure}
\caption{Concept learning curves on small CEBaB datasets with different runs.}
\label{fig:cebab_small_concept_curves}
\end{figure*}

\subsection{Can TBMs work on small training sets?}
\label{sec: Appendix small dataset training}

To evaluate the effect of training on a small dataset, we train three TBMs on the CEBaB dataset after limiting the size of the training set to 50 examples. The learning curves for these TBMs can be seen in figure~\ref{fig:cebab_small_concept_curves}. In this figure, we see that it is possible for TBMs to overfit after generating too many concepts. Across all three runs, we see that performance increases until we reach around 5 concepts, and afterward, it starts to drop. This drop can be explained by the fact that the TBM starts to admit concepts that are too specific, e.g., in the third model, we see a ``Menu Misrepresentation'' concept, which does not exist in any of the five full-size TBM replications. Another explanation for this drop is that there is insufficient information to determine the importance of concepts relative to one another. Thus, even if the correct concepts are generated, the weights assigned to them under low training size samples can be unstable and generalize poorly outside of the training distribution.

\subsection{Learning Curves on all datasets}
\label{sec: Appendix all learning curves}

Figure~\ref{fig:concept_curves_all} shows the concept learning curves on all datasets, in addition to the four reported in Section~\ref{sec: analysis}.

\section{Dataset Details} 
Table \ref{tab:datasets_summary_examples_1} and \ref{tab:datasets_summary_examples_2} show the dataset description, possible labels, and an example from the dataset. Among all datasets, Yelp Reviews, CEBaB, and Hate Speech Detection involve a regression task, while others involve a classification task.

\label{sec: Appendix Datasets}

\begin{table*}[!t]
\centering
\resizebox{\textwidth}{!}{%
\begin{tabular}{>{\raggedright\arraybackslash}p{0.15\linewidth}  p{0.45\linewidth} p{0.4\linewidth}}
\toprule
\textbf{Dataset (Citation)} & \textbf{Description with Possible Labels} & \textbf{Example with Label} \\
\midrule
Fake News Detection \newline \cite{zhong2023goal} & A dataset containing real and fake news from different publishers. \newline \textbf{Possible labels}: Fake, Real & Brazil qualify for 2018 World Cup after Coutinho and Neymar down Paraguay Brazil...  \newline \textbf{Label}: Real  \\
\midrule
Yelp Reviews \newline \cite{zhang2015character} & This dataset contains user-written Yelp reviews. The goal is to predict the review rating (1 to 5 stars) based on the text of the review. \newline \textbf{Possible labels}: 1 Star, 2 Stars, 3 Stars, 4 Stars, 5 Stars & OMG. The best authentic Mexican food. Spicy - yes.  \newline \textbf{Label}: 4 Stars \\
\midrule
Poem Sentiment \newline  \cite{sheng2020investigating} & This dataset contains verses of poems with their sentiment labels. The goal is to predict the sentiment of a verse based on its text. \newline \textbf{Possible labels}: Negative, Positive, No Impact, Mixed & and say, ‘fie, pale-face! are you english girls  \newline \textbf{Label}: No Impact \\
\midrule
Rotten Tomatoes \newline  \cite{Pang+Lee:05a} & This dataset contains movie reviews from Rotten Tomatoes. The goal is to predict the binary sentiment of a review based on its text. \newline \textbf{Possible labels}: Negative, Positive & the performers are so spot on, it is hard to conceive anyone else in their roles.  \newline \textbf{Label}: Positive \\
\midrule
Stanford Natural Language Inference (SNLI) \newline  \cite{snli:emnlp2015} & This dataset contains pairs of sentences (premise and hypothesis). The goal is to predict the relationship between the premise and hypothesis. \newline \textbf{Possible labels}: Entailment, Neutral, Contradiction & Premise: A man giving a speech for the student financial administrators.\newline Hypothesis: There is a man in this picture \newline \textbf{Label}: Neutral \\
\midrule
AG News \newline \cite{Gulli} & A dataset containing news of various categories. The goal is to predict the category based on the news text. \newline \textbf{Possible labels}: Business, Science/Technology, Sports, World/Political & 4 studios back Toshiba HD DVD TOKYO...  \newline \textbf{Label}: Business \\
\midrule
Amazon Reviews \newline \cite{mcauley2013hidden} & This dataset contains product reviews from Amazon. The goal is to predict the binary sentiment of a review based on its text. \newline \textbf{Possible labels}: Negative, Positive & Someone recommended this product to me - it keeps my floors cleaner longer...  \newline \textbf{Label}: Positive \\
\midrule
CEBaB \newline \cite{abraham2022cebab} & Restaurant reviews from OpenTable. \newline \textbf{Possible labels}: 1 Star, 2 Stars, 3 Stars, 4 Stars, 5 Stars & Very poor service and food, this is a second try for this restaurant...  \newline \textbf{Label}: 1 Star \\
\midrule
News Partisanship \newline \cite{kiesel-etal-2019-semeval} & Hyperpartisan News Detection for PAN @ SemEval 2019 Task 4. Given a news article text, decide whether it follows a hyperpartisan argumentation. \newline \textbf{Possible labels}: Not Hyperpartisan, Hyperpartisan & title: Trump Must Now Be Compelled to Withdraw text: This is now bigger than who becomes the next president. Trump is a threat to our democracy.  \newline \textbf{Label}: Hyperpartisan \\
\midrule
Citation Intent \newline \cite{cohan-etal-2019-structural} & A dataset for classifying citation intents in academic papers into method, background, or result. \newline \textbf{Possible labels}: Method, Background, Result & However, the k-safeness of the hypercube does not guarantee the connectivity of the network unless we also bound the number of faulty nodes by 2(n - k) - 1 [17, 35].  \newline \textbf{Label}: Background \\
\bottomrule
\end{tabular}
}
\caption{Summary of Datasets.}
\label{tab:datasets_summary_examples_1}
\vspace{-.2cm}
\end{table*}

\begin{table*}[!t]
\centering
\resizebox{\textwidth}{!}{%
\begin{tabular}{>{\raggedright\arraybackslash}p{0.15\linewidth}  p{0.45\linewidth} p{0.4\linewidth}}
\toprule
\textbf{Dataset (Citation)} & \textbf{Description with Possible Labels} & \textbf{Example with Label} \\
\midrule
Patent Classification \newline \cite{sharma2019bigpatent} & A Patent Classification Dataset classifying patents into various categories. \newline \textbf{Possible labels}: Human Necessities, Performing Operations; Transporting, Chemistry; Metallurgy, Textiles; Paper, Fixed Constructions, Mechanical Engineering; Lightning; Heating; Weapons; Blasting, Physics, Electricity, General tagging of new or cross-sectional technology & a display device has a measuring circuit to detect flicker due to the presence of a dc voltage by monitoring the pixel voltage and, if necessary, modifying driving signals.  \newline \textbf{Label}: Physics \\
\midrule
Hate Speech Detection \newline \cite{kennedy2020constructing} & Online comments with an associated hate speech score. The measure is continuous for hate speech, where higher values indicate more hateful content, and lower values indicate less hateful or supportive speech. A score greater than 0.5 is approximately hate speech, a score less than -1 is counter or supportive speech, and scores between -1 and +0.5 are neutral or ambiguous. \newline \textbf{Possible labels}: Continuous score & I never saw the privilege of being gay until I just saw a guy grab 8 boxes of pregnancy tests at the dollar store \newline \textbf{Label}: -1.48 \\
\bottomrule
\end{tabular}
}
\caption{Summary of Datasets (continued).}
\label{tab:datasets_summary_examples_2}
\vspace{-.2cm}
\end{table*}

\section{Human Evaluation for Concept Generation and Measurement}
\label{sec: appendix human evaluation}

\subsection{Concept Generation}

The authors used the following guide to annotate each concept's quality in Sec~\ref{sec: Appendix concept measurement}. Quality scores equal to 1 indicate no problems, while quality scores greater than 1 indicate issues.

For evaluation, we use the following metrics: (1) \textbf{Redundancy (Rdy)}:
1 - No issues; 
2 - Given the rest of the concepts already generated, this concept is redundant; (2) \textbf{Relevance (Rlv)}:
1 - This concept is related to the task; 
2 - This concept is unrelated to the task; (3) \textbf{Leakage (Lkg)}:
1 - This concept does not leak the labeling task; 
2 - This concept leaks the labeling task; (4)\textbf{Objectivity (Obj)}:
1 - This concept can be measured objectively; 
2 - This concept is subjective. (5) \textbf{Difficulty (Dfc)}:
1 - Answering this concept question is easier than the labeling task; 
2 - Answering this question is around the same difficulty as the labeling task; 
3 - This question is harder than the labeling task.

\subsection{Concept Measurement}
\label{sec: Appendix concept measurement crowd task}
To evaluate the TBM's performance on concept measurement, we generate a questionnaire for each dataset and ask human crowd workers to measure the scores of TBM-generated concepts. To avoid cases where the questions do not have an answer, we insert a ``None of the Above'' response at the end. Our annotators are students from a graduate-level AI class with good English proficiency. Both tasks are given as optional extra credit assignments in the class. Participation is solely voluntary. Before participation, students can preview the tasks and are given a clear description of how the data will be used at the beginning of the instructions. The population size is 98. 

We designed a web interface for the task, which can be found in the Supplementary Materials (Fig~\ref{fig:survey_format}). With an hour of work, students could earn 1\% in extra credit towards their overall course grade.

\section{Additional Details of Implementation}
\label{sec: Appendix implementation}

\subsection{Prompt Structure}

The prompt contains three main sections: The instruction set, dataset information, and TBM state information. The instruction set contains details about the concept generation task, the concept format, and three examples of valid concepts for toxicity detection, product sentiment analysis, and scam detection. This is followed by dataset information which is where we insert the dataset information, label descriptions, and examples from the dataset with different labels. Finally, to avoid making duplicate concepts, we load the list of previously generated concepts at the end in the TBM state information section.

\subsection{Selecting in-context examples}

We selectively load highly misclassified examples during the concept generation stage to increase the chances that we generate concepts that are relevant to these misclassified examples. We get these examples by examining the 10 nearest neighbors for each training example under the current concept feature space and then obtaining 20 examples with the highest ``neighborhood loss'' which is obtained by averaging each neighborhood's MSE (for regression) or accuracy (for classification). We then check to see if this set of examples exceeds the token limit. If it does, we iteratively remove an example with the most common label within the group to ensure diverse representation. If we end up with less than 4 examples, we restart the process but truncate the texts by a factor of 0.8. We note that this sampling is random when the TBM generates its first concept.

\subsection{Managing Token Limits}

The number of iterations we can perform is bounded primarily by the token limit of the LLM that we are using. As the number of iterations increases, the length of the TBM state information grows, and at some point, concept generation fails due to exceeding token limits. It is possible to truncate this step, but that can cause issues with redundant concepts. To help manage token limits in other parts of the prompt, we dynamically truncate the text examples loaded in to ensure that the prompt stays within the token budget of the LLM being used.

\textbf{Concept Measurement}

This module relies heavily on the concept question and response guide associated with each concept to function. This module is flexible, allowing various prompting methods, such as directly answering the question or chain-of-thought prompting. In this paper, we structure the prompt as a three-step process that involves extracting pertinent snippets from the text and reasoning over them before yielding a final answer. The prompt takes in as input the text we want to measure along with the JSON of the concept being evaluated. The prompt returns a JSON object representing the salient snippets in the text for each possible classification, the reasoning of the model over those snippets, and then the final classification. We perform batch inference \cite{cheng2023batch} to reduce LLM costs. In cases where the generated text fails to parse as valid JSON or does not contain text we can turn into a score using the response mapping, we return a concept score of 0. 

\section{Prompts} \label{sec: appendix prompt}

\subsection{Concept Generation Prompt}
\label{sec: appendix concept generation prompt}
\begin{lstlisting}[xleftmargin=.0\textwidth, xrightmargin=.0\textwidth]
Concept Feature Engineering Task

Below we are given a text dataset with accompanying labels. Our task is to identify a concept in the text that could be associated with the label. This is because we want to find the main factors that can be used to explain the label.

To do this, we will examine a sample of texts that have different labels so that we can look at the different characteristics that exist for one label and compare it to another. Good concepts are those that separate texts with one label from another.

After looking at these texts and finding a difference, we will define a concept definition JSON
Each full concept definition comes with a concept name, description, question, response set, and response guide. The concept description provides an intuitive overview of the concept. The concept question is our tool for measuring the concept, this will be graded by a human annotator. The possible responses list the possible responses to the question and the response guide provides information on what each rating means. We also include a response mapping to help with data processing.

Below are some examples of concepts for different datasets.

1. A possible concept for a dataset assigning toxicity scores to social media texts
{"Concept Name": "explicit language",
"Concept Description": "'Explicit language' refers to the use of words, phrases, or expressions that are offensive, vulgar, or inappropriate for general audiences. This may include profanity, obscenities, slurs, sexually explicit or lewd language, and derogatory or discriminatory terms targeted at specific groups or individuals.",
"Concept Question": "What is the nature of the language used in the text?",
"Possible Responses": ["explicit", "strong","non-explicit", "uncertain"],
"Response Guide": {
"explicit": "The text contains explicit language, such as profanity, obscenities, slurs, sexually explicit or lewd language, or derogatory terms targeted at specific groups or individuals.",
"strong": "The text contains strong language but not explicit language, it may contain terms that some viewers might find mature.",
"non-explicit": "The text is free from explicit language and is appropriate for general audiences.",
"uncertain": "It is difficult to determine the nature of the language used in the text or if any explicit terms are used."
},
"Response Mapping": {
"explicit": 2,
"strong":1,
"non-explicit": -1,
"uncertain": 0
}
}###

2. A possible concept for evaluating the sentiment of product reviews on ecommerce site
{"Concept Name": "good build quality",
"Concept Description": "Build quality refers to the craftsmanship, durability, and overall construction of a product. It encompasses aspects such as materials used, design, manufacturing techniques, and attention to detail. A product with good build quality is typically considered to be well-made, sturdy, and long-lasting, while a product with poor build quality may be prone to defects or wear out quickly.",
"Concept Question": "What does the review say about the build quality of the product?",
"Possible Responses": ["positive", "negative", "uncertain", "not applicable"],
"Response Guide": {
"high": "Review mentions aspects such as well-made, sturdy, durable, high-quality materials, excellent craftsmanship, etc.",
"low": "Review mentions aspects such as poor construction, flimsy, cheap materials, bad design, easily breakable, etc.",
"uncertain": "Review does not mention build quality, the information is ambiguous or vague, or it has both positive and negative aspects mentioned like 'the product is sturdy but uses cheap materials'.",
"not applicable": "The review does not mention the build quality of the product at all."
},
"Response Mapping": {
"high": 1,
"low": -1,
"uncertain": 0,
"not applicable": 0
}
}###

3. A useful concept for scam detection for emails
{
"Concept Name": "Extremely generous offer",
"Concept Description": "The concept 'Extremely generous offer' refers to situations where the text describes an offer that seems too good to be true, such as promises of large financial gains, disproportionate rewards, or substantial benefits with seemingly little to no risk or effort required. These can often be indicative of scams or deceptive practices.",
"Concept Question": "What type of offer is described in the text?",
"Possible Responses": ["extremely generous offer", "ordinary offer", "no offer", "uncertain"],
"Response Guide": {
"extremely generous offer": "The text describes an offer that is disproportionately rewarding or beneficial with seemingly little to no risk or effort. This could include promises of large financial returns with minimal investment, 'free' gifts that require payment information, or rewards that are disproportionate to the effort required.",
"ordinary offer": "The text describes a typical or ordinary offer. For instance, normal sales or discounts, standard business offerings, or fair trades.",
"no offer": "The text does not describe any offer.",
"uncertain": "It is difficult to determine the type of offer described in the text. The text might be vague, ambiguous, or lack sufficient context."
},
"Response Mapping": {
"extremely generous offer": 1,
"ordinary offer": -2,
"no offer": -1,
"uncertain": 0
}
}

---

In the task, we will generate concepts for the fake_news dataset

Below is an explanation of the dataset and the labels therein:

Description: A dataset containing real and fake news from different publishers

Label guide: {'0': 'fake', '1': 'real'}

Below are some example texts along with their labels.
---
text:Wells Fargo profits spike despite legal costs "Despite the looming court costs of its recent scandal, Wells Fargo bank has reported an increase in quarterly profits. Third quarter profit rose 2% to $6bn, up from $5.8bn last year. In response, the bank is hiring for positions which were previously cut when employees involved in the scandal were fired. The scandal involved employees opening up fake accounts in customers' names without the customers' knowledge. The latest report of profit increases has surprised many in and outside of the bank."
rating: fake
text:Shailene Woodley to lead hunger strike in jail over pipeline US actress Shailene Woodley has announced to fellow inmates and guards of a North Dakota jail that she will lead a hunger strike against the Dakota Access oil pipeline. Woodley is the start of the Divergent Series. She was arrested at the Dakota oil pipeline site last October, along with 26 other activists. Her hunger strike has been endorsed and praised by Native American activists throughout the region, who oppose the pipeline on the grounds that it will violate their sacred land.
rating: fake
text:Toshiba's Westinghouse files for US bankruptcy Westinghouse Toshiba's US nuclear unit has filed for US bankruptcy protection. The US firm has struggled with hefty losses that have thrown its Japanese parent into a crisis putting the conglomerate's future at risk. Westinghouse has suffered huge cost overruns at two US projects in Georgia and South Carolina. Toshiba said the bankruptcy would not affect Westinghouse's UK operation which employs more than 1 000 workers. However the firm warned that the writedown of its US nuclear business could see Toshiba's total losses last year exceed 1 trillion yen ($9.1bn; 7.3bn) almost triple its previous estimate. The Japanese government confirmed on Wednesday that it was aware of Toshiba's plans.
rating: real
text:George Michael portrait by Damien Hirst sells for $580 000 "A portrait of the late George Michael by artist Damien Hirst has sold for just under half a million pounds at a charity auction. The money raised from the sale of Beautiful Beautiful George Michael Love Painting will go to HIV/Aids charity The Goss-Michael Foundation. The charity was founded by Michael and his former partner Kenny Goss. Goss posted an image of the artwork on Instagram writing: "Amazing result of $580 000 (around 461 011)." He described Damien Hirst as a "superstar" adding: "Thank you Damien!" The canvas went under the hammer in Dallas Texas at the MTV Re:define charity gala. Michael who enjoyed a lucrative pop career as one half of duo Wham before embarking on a successful solo career died on Christmas Day last year at the age of 53.
rating: real
text:Solar-powered 'skin' could make prosthetics more real Many people try to stay out of the sun. But if a new type of solar-powered electronic skin makes its way onto prosthetics, wearers will definitely want those rays shining on their limbs. Researchers are already working to create smart skin that embeds sensors that mimic the tactile feedback of human skin, making it possible for amputees to feel pressure, temperature and even dampness. But how to power the futuristic material? A team from the University of Glasgow in the UK has come up with a version that harnesses the sun's rays. Because it produces its own energy from a natural source, the engineers say, the electronic skin would operate longer than similar materials powered by batteries or tethered to a power source that would also limit portability, clearly a key feature of any everyday prosthetic or touch-sensitive robot on the go.
rating: real
---
As a reminder we already have the following concepts which are useful:
1. Biased language:Biased language refers to the use of words, phrases, or expressions that have an underlying political or ideological agenda. This may include words or phrases that are used to promote a specific point of view, or language that is used to discredit or denigrate certain individuals or groups., possible responses: ['biased', 'non-biased', 'uncertain']
2. Non-credible Sources:Non-credible sources refer to sources of information that lack authority, accuracy, objectivity, or authenticity. This may include sources that are not verified, are not authoritative, or have a poor track record of accuracy., possible responses: ['non-credible source', 'credible source', 'uncertain']
3. Misleading Information:Misleading information refers to statements that are false or misleading, either intentionally or unintentionally. This may include factual errors, inaccurate comparisons, or claims that are not supported by evidence., possible responses: ['misleading', 'accurate', 'uncertain']
4. Exaggerated Claims in News Title:Exaggerated claims in news titles refer to claims made in the title of a news article that exaggerate the truth or lack evidence. This may include statements that are too good to be true, unrealistic promises, or claims that are not supported by evidence., possible responses: ['exaggerated', 'accurate', 'uncertain', 'not applicable']
5. Misleading Language:Misleading language refers to words, phrases, or expressions that are used to mislead or deceive. This may include statements that are false or inaccurate, are presented in a way to distort facts or reality, or are used to manipulate the reader's beliefs., possible responses: ['misleading', 'accurate', 'uncertain']
6. unverified sources:Unverified sources are sources of information which do not have reliable evidence or proof to back them up. This may include sources that are not properly fact checked, are not independently verified, or have a track record of inaccuracy., possible responses: ['unverified source', 'verified source', 'uncertain']

The following concepts have been rejected by the system, avoid making similar ones:
1. exaggerated claims:Exaggerated claims refer to statements that are intentionally exaggerated or hyperbolic in order to create a false sense of urgency or importance. This may include statements that are too good to be true, unrealistic promises, or claims that lack evidence or are not supported by facts., possible responses: ['exaggerated', 'accurate', 'uncertain', 'not applicable']
2. Misleading Headlines:Misleading headlines refer to headlines that are false or misleading, either intentionally or unintentionally. This may include factual errors, inaccurate comparisons, or claims that are not supported by evidence., possible responses: ['misleading', 'accurate', 'uncertain', 'not applicable']
3. Inaccurate Representation:Inaccurate representation refers to statements in the text that are false or misleading, either intentionally or unintentionally. This may include factual errors, inaccurate comparisons, or claims that are not supported by evidence., possible responses: ['accurate', 'inaccurate', 'uncertain']



Keeping in mind the pointers above, create a concept below that is distinct from the current set of concepts. Additionally, make sure that all possible responses can be mapped to an integer.
Make sure that the concept is as relevant to the labels in the fake_news dataset. As the examples shown start to look more similar, we can start being more specific, picking out particular details tied to the label we notice in the examples above.

Definition: {
"Concept Name": "Misleading Information in the Title",
"Concept Description": "Misleading information in the title refers to statements in the title of a news article that are false or misleading, either intentionally or unintentionally. This may include factual errors, inaccurate comparisons, or claims that are not supported by evidence.",
"Concept Question": "What does the title of the article say about the accuracy of the information?",
"Possible Responses": ["misleading","accurate","uncertain","not applicable"],
"Response Guide": {
"misleading": "The title of the article contains misleading information, such as factual errors, inaccurate comparisons, or claims that are not supported by evidence.",
"accurate": "The title of the article contains accurate information.",
"uncertain": "It is difficult to determine the accuracy of the information in the title or if any misleading information is present.",
"not applicable": "The title does not contain any information."
},
"Response Mapping": {
"misleading": 1,
"accurate": -1,
"uncertain": 0,
"not applicable": "na"
}
}

\end{lstlisting}

\vspace{1cm}

\begin{minipage}{\textwidth}
\subsection{Concept Improvement Prompt}
\label{sec: Appendix concept improvement}

\end{minipage}

\begin{lstlisting}[xleftmargin=.0\textwidth, xrightmargin=.0\textwidth]
Concept Improvement Task

We have a concept that needs to be improved. The goal of this task is to identify any issues with the current concept and suggest improvements to make it more valid, clear, well-phrased, and properly formatted in JSON. The concept should be designed for a positive/negative/uncertain question format that maps to 1, -1, and 0 respectively.
In this task, we will return information about any potential problems in the concept along with the improved concept.
Note it is also possible that the concept requires no further improvement (even minor ones), in which case, we will return the original concept with "None" for the other responses.

Consider the following error cases while improving the concept:

1. Lack of validity: Ensure that the responses are mutually exclusive and collectively exhaustive.
- Example of a concept that is not mutually exclusive:
Input JSON:
{{
"Concept Name": "review sentiment",
"Concept Description": "The sentiment expressed towards the product in the review. It could be positive, negative, or neutral.",
"Concept Question": "What is the overall feeling towards the product?",
"Possible Responses": ["positive", "somewhat positive", "negative"],
"Response Guide": {{
"positive": "The reviewer expresses a positive opinion on the product, such as praising its quality, performance, or value.",
"somewhat positive": "The reviewer expresses a somewhat positive opinion on the product, such as mentioning some good aspects but also pointing out some flaws.",
"negative": "The reviewer expresses a negative opinion on the product, such as criticizing its quality, performance, or value."
}},
"Response Mapping": {{
"positive": 1,
"somewhat positive": 0.5,
"negative": -1
}}
}}
Response: {{
"Confirmation":"1. Mutual Exclusivity: The concept above contains non-mutually exclusive responses 'positive' and 'somewhat positive'
2. Collectively Exhaustive: clear
3. No leading questions: clear
4. Rich and objective response guide: clear
5. Interference with other concepts: clear
6. Invalid response set: clear",
"Errors": "The concept above contains non-mutually exclusive responses 'positive' and 'somewhat positive'",
"Fix": "We can address this by either combining 'positive' and 'somewhat positive' into a single response or defining clearer distinctions between them.",
"New Concept": {{
"Concept Name": "review sentiment",
"Concept Description": "The sentiment expressed towards the product in the review. It could be positive, negative, or neutral.",
"Concept Question": "What is the overall sentiment expressed towards the product in the review?",
"Possible Responses": ["positive", "negative", "neutral"],
"Response Guide": {{
"positive": "The reviewer expresses a positive opinion on the product, such as praising its quality, performance, or value.",
"negative": "The reviewer expresses a negative opinion on the product, such as criticizing its quality, performance, or value.",
"neutral": "The reviewer does not express a clear positive or negative opinion on the product, or the review contains a mix of positive and negative aspects."
}},
"Response Mapping": {{
"positive": 1,
"negative": -1,
"neutral": 0
}}
}}
}}###

- Example of a concept that is not collectively exhaustive:
Input JSON:
{{
"Concept Name": "product availability",
"Concept Description": "The availability of the product as described in the review.",
"Concept Question": "Is the product available?",
"Possible Responses": ["available", "unavailable"],
"Response Guide": {{
"available": "The reviewer mentions that the product is available, in stock, or easy to find.",
"unavailable": "The reviewer mentions that the product is unavailable, out of stock, or hard to find."
}},
"Response Mapping": {{
"available": 1,
"unavailable": -1
}}
}}
Response: {{
"Confirmation": "Confirmation:
1. Mutual Exclusivity: clear
2. Collectively Exhaustive: The concept above contains a non-collectively exhaustive response set
3. No leading questions: clear
4. Rich and objective response guide: clear
5. Interference with other concepts: clear
6. Invalid response set: clear"
"Errors": "The concept above contains a non-collectively exhaustive response set because it may be possible that a piece of text does not strictly match some criteria in the response guide,",
"Fix": "We can address this by adding a 'uncertain' response to cover cases where the availability is not clearly mentioned, and a 'not applicable' response for cases where the text does not discuss a product.",
"New Concept": {{
"Concept Name": "product availability",
"Concept Description": "The availability of the product as described in the review.",
"Concept Question": "What does the review say about the product's availability?",
"Possible Responses": ["available", "unavailable", "uncertain", "not applicable"],
"Response Guide": {{
"available": "The reviewer mentions that the product is available, in stock, or easy to find.",
"unavailable": "The reviewer mentions that the product is unavailable, out of stock, or hard to find.",
"uncertain": "The review contains mixed information or no information that makes it difficult to determine the availability of the product.",
"not applicable": "The reviewer is not discussing a product or anything else that could be described by this concept."
}},
"Response Mapping": {{
"available": 1,
"unavailable": -1,
"uncertain": 0,
"not applicable": "na"
}}
}}
}}###

2. Poor phrasing: Avoid leading questions and provide rich examples in the response guide.
- Example of a leading question:
Input JSON:
{{
"Concept Name": "product cost",
"Concept Description": "The cost of the product as described in the review.",
"Concept Question": "Is the product expensive?",
"Possible Responses": ["expensive", "affordable", "uncertain", "not applicable"],
"Response Guide": {{
"expensive": "The reviewer thinks the product is expensive.",
"affordable": "The reviewer thinks the product is affordable.",
"uncertain": "The reviewer does not provide a clear opinion on the product's cost.",
"not applicable": "The reviewer is not discussing a product or anything else that could be described by this concept."
}},
"Response Mapping": {{
"expensive": 1,
"affordable": -1,
"uncertain": 0,
"not applicable": "na"
}}
}}
Response: {{
"Confirmation":"1. Mutual Exclusivity: clear
2. Collectively Exhaustive: clear
3. No leading questions: The question may potentially be leading, "Is the product expensive?" leads the responder to a positive response.
4. Rich and objective response guide: The response guide is not very rich or objective. The responses guides for each answer are very similar, only differing by a few words and without providing examples.
5. Interference with other concepts: clear
6. Invalid response set: clear",
"Errors": "The concept contains a leading question and a bad response guide",
"Fix": "We can address this by changing the question to be more neutral and asking about the reviewer's description of the product's cost.",
"New Concept": {{
"Concept Name": "product cost",
"Concept Description": "The cost of the product as described in the review, in terms of whether the product is perceived as expensive or affordable.",
"Concept Question": "How does the reviewer describe the cost of the product?",
"Possible Responses": ["expensive", "affordable", "uncertain", "not applicable"],
"Response Guide": {{
"expensive": "The reviewer describes the product as costly, high-priced, or not worth the money.",
"affordable": "The reviewer describes the product as reasonably priced, good value for money, or budget-friendly.",
"uncertain": "The reviewer does not provide a clear opinion or information on the product's cost.",
"not applicable": "The reviewer is not discussing a product or anything else that could be described by this concept."
}},
"Response Mapping": {{
"expensive": 1,
"affordable": -1,
"uncertain": 0,
"not applicable": "na"
}}
}}
}}###

In addition to the errors above, some other problems could be:

1. The concept contains responses not in {{1, 0, -1, "na"}}. This can be fixed by shrinking the possible set of responses.
2. Lack of detail in the response guide. As much as possible, the response guide should contain detailed examples. This issue can be fixed by making the response guide more specific to allow annotators to be more objective about answering the question.

---
Below is the concept for you to improve.
{}
Response:{{

\end{lstlisting}

\subsection{Concept Measurement Prompt} \label{sec: Appendix concept measurement}

\begin{lstlisting}[xleftmargin=.0\textwidth, xrightmargin=.0\textwidth]

Answer the following question about the texts below by selecting from the following choices. Before answering the question, extract any potentially relevant snippets of the text that can serve as evidence for each classification. After that, compare the snippets against the response guide to come up with a final decision.

Format your response as a list of JSON objects with string keys and string values. Below is an example of a valid JSON response. Each JSON object contains keys for snippets, thoughts, and answer. End your response with ###
---
Text 1: Text
Text 2: Text
Text 3: Text

Response JSON:[
{"text": "Text 1", "snippets": {
"classification 1" : ["Snippet 1", "Snippet 2", ...],
"classification 2" : ["Snippet 3", "Snippet 4", ...]
...
},
"thoughts": "In this section, you weigh evidence based on the text and the extracted snippets to come to a final decision with the response guide as a reference. Be as objective as possible and ignore irrelevant information. Focus only on the snippets and avoid making guesses.",
"answer": "An answer from the response guide goes here. In answering the question, ignore irrelevant information and avoid making assumptions."},
{"text": "Text 2", "snippets": {
"classification 1" : ["Snippet 1", "Snippet 2", ...],
"classification 2" : ["Snippet 3", "Snippet 4", ...]
...
},
"thoughts": "...",
"answer": "..."},
{"text": "Text 3", "snippets": {
"classification 1" : ["Snippet 1", "Snippet 2", ...],
"classification 2" : ["Snippet 3", "Snippet 4", ...]
...
}
]###
---
Below is an example of the task being performed with the concept "build quality":

Concept: 
{
"Concept Name": "good build quality",
"Concept Description": "build quality refers to the craftsmanship, durability, and overall construction of a product. It encompasses aspects such as materials used, design, manufacturing techniques, and attention to detail. A product with good build quality is typically considered to be well-made, sturdy, and long-lasting, while a product with poor build quality may be prone to defects or wear out quickly.",
"Concept Question": "What does the review say about the build quality of the product?",
"Possible Responses": ["high", "low", "uncertain", "not applicable"],
"Response Guide": {
"high": "Review mentions aspects such as well-made, sturdy, durable, high-quality materials, excellent craftsmanship, etc.",
"low": "Review mentions aspects such as poor construction, flimsy, cheap materials, bad design, easily breakable, etc.",
"uncertain": "Review does not mention build quality, the information is ambiguous or vague, or it has both positive and negative aspects mentioned like 'the product is sturdy but uses cheap materials'.",
"not applicable": "The review does not mention the build quality of the product at all."
}
}

Text 1: "This product has a great design and is really easy to use. It is also very durable."
Text 2: "Was excited for it to finally arrive, got here in nice sturdy packaging. Opened it up though and it smelled kind of weird? goes away after a while but otherwise an ok product. Saw some print aberrations it didn't interfere much with use."
Text 3: "A big fan of the product. Serves me well during workouts but I go through them like hotcakes. Don't expect it to last long compared to other brands but you get what you pay for. It does the job though."
Text 4: "Very disappointing. I was excited to order this but when it arrived I was shocked at how poorly it worked. Deceptive advertising at its finest."

Response JSON:[
{"text": "Text 1", "snippets": {
"high": ["It is also very durable", "really easy to use"],
"low": [],
"uncertain": []
},
"thoughts": "Two snippets for high. The first is related to durability which is an aspect of good build quality. The second is related to ease of use which is not related to good build quality. Overall the text describes good build quality..",
"answer": "high"},

{"text": "Text 2", "snippets": {
"high": ["got here in nice sturdy packaging"],
"low": ["Saw some print aberrations"],
"uncertain": []
},
"thoughts": "One snippet for high, one snippet for low. The low snippet mentions defects in manufacturing. In the high snippet, 'Sturdy' only refers to the packaging, not the product. The balance of evidence leans towards a classification of low.",
"answer": "low"},

{"text": "Text 3", "snippets": 
{
"high": ["Serves me well during workouts"],
"low": ["Don't expect it to last long compared to other brands", "I go through them like hotcakes"],
"uncertain": ["you get what you pay for"]
},
"thoughts": "One high snippet, two low snippets, one uncertain snippet. The high snippet is about utility which is not related to build quality. The low snippet relates to durability, an aspect of build quality.",
"answer": "low"},

{"text": "Text 4", "snippets":
{
"high": [],
"low": ["Very disappointing", "shocked at how poorly it worked", "Deceptive advertising"],
"uncertain": []
},
"thoughts": "Three low spans. The first is related to overall judgment which is irrelevant, the second is related to functionality which is irrelevant, and the third is related to marketing/advertising which is also irrelevant. None are related to build quality.",
"answer": "uncertain"
}
]###
---
Perform the task below, keeping in mind to limit snippets to 10 words and ignoring irrelevant information. Return a valid list of JSON objects ending with ###
Concept: {'Concept Name': 'Price to Quality Ratio', 'Concept Description': "Price to Quality Ratio is a measure of a customer's perception of the value of the goods or services they receive relative to the price they paid. A high price to quality ratio indicates that the customer believes they have received good value for their money, while a low price to quality ratio suggests that the customer believes they did not receive enough in return for the money they paid.", 'Concept Question': "What is the customer's perception of the price to quality ratio of the goods or services they received?", 'Response Guide': {'high': 'The customer believes they have received a good value for their money, such as feeling that the product is worth more than what they paid for it.', 'average': 'The customer believes they received an average value for their money, such as feeling that the product is worth the same as what they paid for it.', 'low': 'The customer believes they did not receive enough in return for the money they paid, such as feeling that the product is not worth what they paid for it.', 'uncertain': "It is difficult to determine the customer's perception of the price to quality ratio, such as when the customer does not provide a clear opinion or information on the product's cost."}}

Text 1: I'm very very torn about how many stars to give Plum. Very torn.\n\nHere are the positive things:\n+great sushi\n+great interior\n+very solid appetizers and a knock-out hot and sour soup\n+good drink menu\n\nBut the negatives, oh...\n-service is almost non-existent. I've had to get up after 15 minutes of being ignored after being seated to ask if we have a waiter. No one even brought us water (!!) Then, the woman I asked, instead of apologizing, made me feel guilty by telling me that she wasn't my waiter, but she supposed she'd get us water and take our order (I suspect she was our waiter). This is not the only service horror story I have\n-price. Look, I get that this place is nice, but you can't charge this much for food and have service this bad. It's just insulting to have a waitress in sneakers when you're paying for a $30 plate of sushi. \nOh, and more staff issues. They have some guy busing tables in a baseball hat who has certainly been stoned every time I've been there. It's amazing. \n\nFood: pretty great, if a little pricey. \nService, staff, etc: Horrifically bad. \n\nFire the staff, re-hire people who know how to dress appropriately for the atmosphere, politely attend to customers, etc. Also, if you're going to have non-wait staff bus tables, you need to have them dress as waitstaff and politely take items. The guy they have now is like a (stoned) bull in a china shop. No one wants to pay that amount of money for food and be ignored.
Text 2: The light rail has become a big part of my life.  I moved into a place specifically for it's proximity to the Thomas/Central station.  I go to ASU and have ridden it every day this week, here's why:\n\n1. It's free, because I'm an ASU student\n2. It makes the walk at 6:30am in the dark seem kinda cool\n3. I enjoy sitting and not driving\n4. I finished a book with just my metro reading time\n5. I get some much needed \"me\" time\n6. I never have to deal with ASU's parking EVER AGAIN\n\nThanks light rail for keeping me sane and helping save me money until my goddamn FAFSA check comes in the mail.  I'm taking off a star because you don't run late enough to take back from work.\n\nI know you're thinking \"jesus, it sounds like this girl's life sucks hard core\".  Yeah, it really does.

Response JSON:[
{{
"text": "Text 1", "snippets": {
"high": ["great sushi", "great interior", "very solid appetizers", "knock-out hot and sour soup", "good drink menu"],
"low": ["price", "service is almost non-existent", "no one even brought us water", "made me feel guilty", "can't charge this much for food and have service this bad", "non-wait staff bus tables"],
"average": ["Food: pretty great, if a little pricey"],
"uncertain": []
},
"thoughts": "The text contains five snippets for high, six snippets for low, and one snippet for average. The snippets for high are related to the quality of the goods or services, while the snippets for low are related to the price and service. The balance of evidence leans towards a classification of low.",
"answer": "low"
},

{"text": "Text 2", "snippets": {
"high": ["It's free", "makes the walk at 6:30am in the dark seem kinda cool", "I enjoy sitting and not driving", "I finished a book with just my metro reading time", "I get some much needed \"me\" time", "I never have to deal with ASU's parking EVER AGAIN"],
"low": ["doesn't run late enough to take back from work"],
"average": [],
"uncertain": []
},
"thoughts": "The text contains six snippets for high and one snippet for low. The snippets for high are related to the quality of the goods or services, while the snippet for low is related to the price. The balance of evidence leans towards a classification of high.",
"answer": "high"
}]

\end{lstlisting}

\clearpage
\begin{figure*}
\begin{center}
\includegraphics[width=12cm]{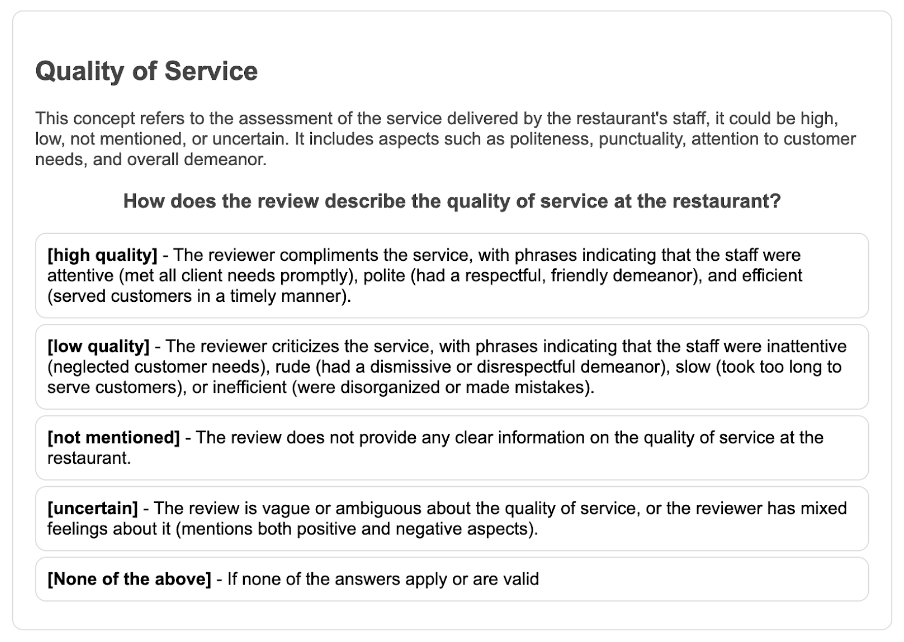}
\end{center}
\caption{Survey format. Annotators for concept measurement answer questions such as the one above for texts in the CEBaB dataset.}
\label{fig:survey_format}
\end{figure*}

\section{Supplementary Material}

Figure~\ref{fig:survey_format}  shows an example of a survey question used to measure the concept ``Quality of Service''.

\end{document}